\documentclass[11pt]{article}
\usepackage[T1]{fontenc}
\usepackage[utf8]{inputenc}
\usepackage{lmodern}
\usepackage[a4paper,margin=26mm]{geometry}
\usepackage{amsmath,amssymb,mathtools}
\usepackage{graphicx,booktabs,tabularx,array}
\usepackage{caption}
\usepackage{xcolor,microtype}
\usepackage{colortbl}
\usepackage[numbers,sort&compress]{natbib}
\usepackage{tikz}
\usetikzlibrary{positioning,arrows.meta}
\usepackage[colorlinks=true,allcolors=blue!50!black]{hyperref}
\usepackage{xurl}
\newcommand{\KV}{\textsc{KV}}
\newcommand{\KVctx}{\textsc{KV-Ctx}}
\newcommand{\KVdec}{\textsc{KV-Dec}}
\DeclareMathOperator{\tr}{tr}

\title{Post-hoc Epistemic Uncertainty for\\ Vision--Language--Action Policies}
\author{%
  Johannes Keller \quad Albrecht M. Bloße \quad Adrian Lindenmeyer\\
  Thomas Neumuth \quad Daniel Schneider\\[1ex]
  \normalsize Innovation Center Computer Assisted Surgery (ICCAS)\\
  \normalsize Leipzig University, Leipzig, Germany%
}
\date{\today}

\begin{document}
\maketitle
\begin{abstract}

Vision–language–action (VLA) policies need to recognize gaps in their training knowledge to request useful demonstrations or assistance in new situations, especially when failures are costly.
Disagreement within an ensemble of independently trained policies can guide these decisions, but requires training and running multiple policies.
We propose post-hoc kernel-variance heads that estimate epistemic uncertainty from the embeddings of a single policy.
Each head computes the conditional variance of a Gaussian process conditioned on embeddings from the training demonstrations.
Fitting requires no regression targets or gradient updates and leaves the policy unchanged.
In experiments with SmolVLA on LIBERO-10, the kernel variance heads perform competitively with two-policy ensemble scores in task-level correlation with task success, in demonstration acquisition during active learning, and in referral to a stronger supervisor policy as a proxy for human assistance.
For SmolVLA, the head on the vision–language backbone reduces uncertainty-scoring time by up to approximately a factor of 50 compared with ensemble scores, while also substantially reducing setup costs.
We also show that the established evaluation protocol on LIBERO-10 primarily rewards recognition of coverage differences between tasks, while execution variability weakens assessment of within-task rankings.
Thus, controlled perturbations provide a complementary evaluation: modality-specific heads identify unfamiliar camera, instruction, or state inputs, while fused heads can detect unfamiliar combinations of individually familiar inputs.
With a supervisor proxy trained on the evaluated perturbation families, the decoder head improves within-task referral over ensemble disagreement.
These findings show that useful uncertainty estimates for adaptive robot learning can be obtained from a single policy.
Code available at \url{https://github.com/post-hoc-vla-uncertainty/post-hoc-vla-uncertainty}.

\end{abstract}
\section{Introduction}
\label{sec:introduction}
Vision--language--action (VLA) models bring pre-trained visual and linguistic representations to robot control~\citep{brohan2023rt1,zitkovich2023rt2,octo2024,kim2024openvla,black2024pi0,black2025pi05,shukor2025smolvla}.
After deployment, such a policy keeps encountering tasks, scenes and instructions that its training demonstrations cover only partially.
In tasks where failures may be costly, when facing difficulties due to insufficient training knowledge, the robot should respond adaptively by deferring tasks to seek human assistance or by requesting relevant additional training demonstrations~\citep{ross2011dagger,kelly2019hgdagger,hoque2021thriftydagger,hoque2022fleetdagger,liu2023sirius,bagatella2025active,romer2026uqvla}.
Both responses are effective only if guided by the policy's epistemic uncertainty, that is, by how poorly its training data support a situation.
Unlike aleatoric ambiguity, epistemic uncertainty shrinks with more relevant demonstrations~\citep{kendall2017uncertainties,hullermeier2021aleatoric}.

Flow-based VLAs do not provide such an estimate by themselves.
This motivates dedicated methods for quantifying epistemic uncertainty in these policies.
For example, \citet{romer2026uqvla} derive velocity-field disagreement (VFD), an ensemble estimate of the divergence between flow-matching policies.
They use VFD in SAVE, an active fine-tuning procedure that they evaluate on LIBERO~\citep{liu2023libero}, a simulation benchmark of language-instructed household manipulation tasks.
In each round, SAVE requests a few new demonstrations, preferring situations that VFD rates as uncertain, and then fine-tunes the policy.
In their evaluation, VFD correlates more strongly with task success than six alternative uncertainty scores, including other ensemble-based measures and entropy-based scores.
SAVE with VFD needs at least 22\% fewer demonstrations than the compared acquisition strategies.
The price is an ensemble of at least two policies: every additional member must be pre-trained, fine-tuned alongside the deployed policy in every adaptation round, and integrated jointly with it whenever a score is computed.

We ask whether this ensemble is necessary to obtain comparably useful estimates of epistemic uncertainty.
Ensemble disagreement reflects differences between the predictions of independently trained policies, which can arise where the demonstrations leave their behaviour weakly constrained.
Their disagreement can therefore serve as an indirect signal of demonstration coverage.
This motivates assessing coverage directly in the embedding space of a single trained policy, using lightweight uncertainty estimation heads.
Each head reads embeddings that the policy already computes and compares a queried observation with support observations from its training demonstrations (Figure~\ref{fig:architecture}).
The approach assumes that the embeddings keep unfamiliar observations apart from familiar ones~\citep{vanamersfoort2021due}, a property that we do not enforce during training.
Its practical usefulness further depends on whether unfamiliarity helps identify situations where the policy may fail or benefit from assistance or additional demonstrations.
We assess these relationships empirically across our experiments.
The measure is the posterior variance of a Gaussian process (GP) with a Gaussian kernel, conditioned on the support embeddings~\citep{rasmussen2006gpml}.
It needs no regression targets and no gradient update and does not alter the policy's predicted actions but can inform decisions such as when to request help or new demonstrations.
Fitting it post hoc to a trained policy takes one feature extraction pass over the support observations and one Cholesky factorization.
Since the heads are this cheap, several of them can be attached to different embeddings of the same policy.
Heads on modality-specific representations, such as the camera or instruction tokens, can then indicate \emph{which} part of an input is unfamiliar.

We organize the study around five questions.
\begin{itemize}
\item[\textbf{Q1:}] Does a single-policy kernel variance head match ensemble scores in calibration to task success and referral gain to a stronger supervisor on LIBERO-10?\footnote{Here, referral means handing the most uncertain situations to a more capable supervisor policy, which serves as a proxy for human assistance.}
\item[\textbf{Q2:}] Does active fine-tuning with the SAVE protocol guided by kernel variance scores acquire demonstrations as effectively as ensemble scores?
\item[\textbf{Q3:}] How much computation does the kernel variance head save compared to ensemble scores?
\item[\textbf{Q4:}] Which properties of an uncertainty score do the established evaluations of Q1 reward on LIBERO?
\item[\textbf{Q5:}] Do kernel variance heads on different context embeddings respond selectively to perturbations of the inputs they read, and do kernel variance heads support within-task referral under such perturbations?
\end{itemize}
In addressing these questions, we make three contributions:
\begin{itemize}
\item Post-hoc epistemic uncertainty heads that read the embeddings of a single trained VLA policy.
Instead of pre-training, fine-tuning and running a second policy, they require only one feature extraction pass over the support observations and one matrix factorization. In our experiments, this makes the heads up to about 44 times faster to set up per fine-tuning round and up to about 50 times faster to use.
\item A comparison on the LIBERO manipulation benchmark, in which the kernel variance heads perform comparably to uncertainty scores from a two-policy ensemble in ranking tasks by their failure rate and in selecting new training demonstrations, following the setting of prior ensemble-based work~\citep{romer2026uqvla}.
We additionally evaluate success gain from handing uncertain situations over to a stronger policy, where the kernel variance heads again match or exceed the ensemble.
Our analysis shows that in this setting, the evaluations mainly test whether a score recognizes how well training demonstrations cover a task, which is useful in itself.
However, they test the ranking of situations within a task only weakly, because relatively large execution noise obscures reproducible differences in the outcomes.
\item A within-task evaluation that perturbs single input modalities and their combinations in a controlled way.
In this evaluation, our uncertainty heads detect most perturbations at least as sensitively as the ensemble scores and match or exceed them in deciding which situations within a task to hand over to a stronger policy.
Attached to the embeddings of separate input modalities, context-specific uncertainty heads also indicate which of them are unfamiliar.
\end{itemize}
Section~\ref{sec:related} places the heads among distance-aware uncertainty estimators, uncertainty scores for generative policies, and work on active data acquisition and deferral.
Section~\ref{sec:methods} defines the kernel-variance heads on SmolVLA and their context-specific variants, the SAVE acquisition loop, the calibration and referral metrics, our matched experimental protocol and the controlled perturbations of single input modalities.
Section~\ref{sec:results} first compares the heads with the ensemble scores in task-level calibration, demonstration acquisition and its cost, and supervisor referral (Q1--Q3).
It then examines which properties of a score the established evaluations reward (Q4) and evaluates the context-specific and whole-policy heads under controlled perturbations (Q5).
Section~\ref{sec:discussion} discusses the five research questions in turn, then further findings, such as why unfamiliarity does not imply failure, and the limitations of our study.

\section{Related Work}
\label{sec:related}

\paragraph{VLA policies and their evaluation.}

VLAs often fine-tune vision-language backbones on large robot datasets~\citep{oneill2024oxe} and output actions as discrete tokens~\citep{brohan2023rt1,zitkovich2023rt2,kim2024openvla} or through continuous generative heads~\citep{octo2024,black2024pi0,black2025pi05,shukor2025smolvla}.
Flow-matching action experts~\citep{lipman2023flow}, like diffusion policies~\citep{chi2023diffusion}, sample action chunks from multimodal distributions.
On LIBERO~\citep{liu2023libero}, a widely used simulation benchmark for these models, high scores may reflect memorized layouts and instructions and do not necessarily imply robustness under changed object positions, cameras, lighting or wording~\citep{zhou2025liberopro,fei2025liberoplus}.
This motivates evaluations that vary conditions \emph{within} a task (Sections~\ref{sec:results-benchmark}--\ref{sec:results-perturbed}).

\paragraph{Epistemic uncertainty and distance awareness.}

Epistemic uncertainty reflects limited knowledge about the true function, whereas aleatoric uncertainty reflects inherent randomness in the data and cannot be reduced by collecting more~\citep{kendall2017uncertainties,hullermeier2021aleatoric}.
For expressive models, epistemic uncertainty is mainly uncertainty about which of the many functions that fit the data is correct, so it is large where data sparsity leaves predictions weakly constrained ~\citep{hullermeier2021aleatoric}.
For a robot policy, it concerns which actions the training demonstrations imply in a given situation, while aleatoric uncertainty arises, for example, when demonstrators solve a task in several valid ways or when contact outcomes are stochastic.
A principled estimate may approximate the posterior over functions and measures the spread of its predictions~\citep{blattmann2025implicit}.

Weight-space methods such as deep ensembles~\citep{lakshminarayanan2017ensembles}, Monte Carlo dropout~\citep{gal2016dropout} and Laplace approximations~\citep{kristiadi2020bayesian,daxberger2021laplace} define the prior over functions only implicitly through the weights.
In practice, ensemble members and weight-space particles often converge to similar functions~\citep{abe2023pathologies,dangelo2021stein,dangelo2021repulsive}, mean-field variational approximations can underestimate the variance between separated data regions~\citep{foong2019inbetween}, and the resulting spread can follow aleatoric noise more closely than data density~\citep{lindenmeyer2024inadequacy,blattmann2025implicit}.
Depending on protocol, decisions such as referral and acquisition may only need a ranking, in which case an underestimated functional spread can still serve them if its ordering is right, although this is not guaranteed.
Even single-model entropy or confidence scores can appear to capture epistemic uncertainty~\citep{hendrycks2017baseline,pearce2021softmax}.
However, by design, they cannot capture the uncertainty associated with their single function.
Implicit biases of trained networks~\citep{pearce2021softmax} or aleatoric estimates that become unreliable away from the training data~\citep{valdenegro2022deeper} can nevertheless make these scores correlate with training-data coverage.
Such agreement is indirect and does not mean that the scores represent epistemic uncertainty, either separately or as an explicit component of total uncertainty.

By contrast, methods with explicit priors over functions allow predictive variance to grow with distance from the training data, with the predictive distribution reverting to the prior far from training support. Examples include GP output layers on spectrally normalized networks~\citep{liu2020sngp,vanamersfoort2021due}, deep GPs~\citep{damianou2013deepgp} and functional variational Bayesian neural networks~\citep{sun2019functional}.
Deterministic scores based on radial basis function prototypes, feature densities, Mahalanobis distance or nearest neighbours also use feature-space distances or densities, but do not generally define a posterior over predictive functions ~\citep{vanamersfoort2020duq,mukhoti2023ddu,lee2018mahalanobis,sun2022knn}.
In direct comparisons, such distance-aware single models match or improve upon deep ensembles in detecting unfamiliar inputs~\citep{vanamersfoort2020duq,mukhoti2023ddu,lindenmeyer2025trustworthy,blattmann2025implicit}.
Methods that score distances in learned representations depend on a feature space that keeps unfamiliar inputs apart from familiar ones, which feature collapse can prevent~\citep{vanamersfoort2021due}.

Our uncertainty scoring head places such an explicit GP prior over functions defined on embeddings of a frozen policy, post hoc.
Spectral normalization could make these embeddings more distance-preserving, but introducing it would modify the policy and would require further retraining.
Instead, we leave the policy unchanged so that the head can be attached to any trained policy.
The head's score, the GP conditional variance~\citep{rasmussen2006gpml}, vanishes at support inputs and approaches the prior variance far from them.
Without the ridge, its square root equals the power function of kernel interpolation, which is the worst-case interpolation error over the unit sphere of the kernel's reproducing kernel Hilbert space~\citep{wendland2004scattered,kanagawa2018gaussian}.
It therefore measures coverage in the chosen embedding geometry without requiring output targets to fit the head.

\paragraph{Uncertainty for generative policies and VLAs.}

For diffusion and flow models, epistemic uncertainty has been estimated from ensemble disagreement of denoising means (DECU)~\citep{berry2024decu}, from a Gaussian approximation of the generative posterior predictive (GU)~\citep{jazbec2025gu}, and from velocity-field disagreement (VFD)~\citep{romer2026uqvla}.
Runtime monitors for robot policies use action-chunk entropy and embedding novelty~\citep{romer2025fiper}, temporal action consistency~\citep{agia2024stac}, learned density estimators~\citep{xu2025faildetect} or language-model embeddings~\citep{sinha2024aesop}.
VLA failure detectors learn from internal features~\citep{gu2025safe}, combine last-layer Mahalanobis distance with action consistency~\citep{seligmann2026vlafail}, or measure action disagreement under hidden-activation perturbations of a single model~\citep{lee2026pfd}, and VLA confidence has been calibrated with prompt ensembles and Platt scaling~\citep{zollo2025calibration}.
Most of these robotics methods focus on failure detection during execution.

Outside robotics, attribution methods explain an existing uncertainty estimate post hoc by its input features, through counterfactuals~\citep{antoran2021clue,perez2022attribution}, gradients~\citep{wang2023uabackprop} or Shapley values~\citep{watson2023shapley}.
Multimodal models estimate uncertainty per modality to weight their fusion~\citep{subedar2019audiovisual,han2021trusted} and use disagreement between modalities for OOD detection~\citep{dong2024multiood}, and a recent benchmark uses modality-specific and cross-modal perturbations to test how vision–language uncertainty scores respond to different sources of uncertainty~\citep{wang2026vlmuqbench}.
Our raw context kernel-variance heads assess unfamiliarity separately for each input modality by reading its representation before cross-modal mixing.
To our knowledge, this is the first use of separate coverage estimates to assess unfamiliarity in the individual input modalities of a VLA policy.
Our heads can inform decisions taken before execution, such as requesting additional training demonstrations or human assistance, and can also score observations during execution.

\paragraph{Active learning and interactive imitation.}

Early work on data selection by expected information gain or variance reduction includes~\citet{mackay1992information} and~\citet{cohn1996active}.
Deep active learning uses Bayesian disagreement~\citep{houlsby2011bald,gal2017deepactive,kirsch2019batchbald} or representation diversity~\citep{sener2018coreset,ash2020badge}, or combines uncertainty and diversity in gradient space~\citep{ash2020badge}.
GP methods use integrated variance reduction and mutual information~\citep{seo2000gp,krause2008sensor}, and transductive active learning targets a specified population~\citep{hubotter2024transductive}.
Interactive imitation learning collects expert feedback on situations encountered by the learner~\citep{ross2011dagger,kelly2019hgdagger}, with some methods using uncertainty or risk to decide when to query the expert~\citep{zhang2017safedagger,menda2019ensembledagger,cui2019uncertainty,hoque2021thriftydagger}.
Related approaches address supervision across robot fleets \citep{hoque2022fleetdagger} and learning during deployment \citep{liu2023sirius}.
Data curation selects which existing demonstrations to keep~\citep{hejna2025curation}, whereas multi-task active fine-tuning selects which new demonstrations to collect~\citep{bagatella2025active,romer2026uqvla}.
Of these, we build on the SAVE procedure of~\citet{romer2026uqvla}, keep its rule for choosing demonstrations and replace only its uncertainty score, VFD, with the variance of our kernel heads.

\paragraph{Selective prediction and deferral.}

Selective prediction allows a model to abstain on uncertain inputs and evaluates uncertainty scores through the trade-off between coverage (the fraction of inputs retained) and risk on the retained inputs~\citep{chow1970optimum,elyaniv2010foundations,geifman2017selective,leibig2017leveraging,band2021retina}.
This trade-off does not distinguish epistemic from aleatoric uncertainty, since rejecting inherently ambiguous inputs also lowers retained risk.
Combined with the fact that many scores labelled epistemic remain entangled with aleatoric noise~\citep{valdenegro2022deeper,dejong2024disentangled,lindenmeyer2024inadequacy,blattmann2025implicit}, this creates a major evaluation limitation: improved performance can reflect sensitivity to inherent ambiguity rather than the identification of missing knowledge.
Deferral passes rejected inputs to a second decision maker, whose errors also affect combined performance.
Some evaluations of combined performance assume that an oracle correctly labels every deferred case~\citep{kivlichan2021moderator} and therefore inherit the same limitation.
Learning to defer formulates the delegation decision as a learning problem~\citep{madras2018defer,mozannar2020defer} and has also been studied with imperfect simulated experts~\citep{hemmer2023defer}.
Deferral also arises in robotics, where calibrated uncertainty guides requests for assistance~\citep{ren2023knowno,angelopoulos2023conformal}.
Our supervisor referral follows the deferral view with an imperfect expert.
Under a fixed referral budget, we pass selected conditions to a stronger policy and evaluate each ranking by the supervisor's actual, possibly negative, performance gain on the referred conditions.
Where irreducible randomness affects both policies alike, referring a condition contributes no expected gain.
This reduces, but not eliminates, the credit a score receives for detecting ambiguity alone.
The metric primarily rewards identifying conditions for which the supervisor's broader training coverage provides a performance advantage.

\section{Methods}
\label{sec:methods}
\subsection{Kernel-variance heads}
\label{sec:kv}
\begin{figure}[t]
\centering
\begin{tikzpicture}[node distance=4mm and 9mm, >=Latex,
 block/.style={draw=black!55,rounded corners=2pt,fill=black!3,align=center,font=\small,text width=62mm,minimum height=7mm},
 tap/.style={draw=blue!55!black,fill=blue!5,rounded corners=2pt,align=left,font=\scriptsize,text width=56mm,minimum height=7mm},
 ctx/.style={tap,draw=orange!75!black,fill=orange!7},
 read/.style={->,dashed,blue!55!black},
 readctx/.style={->,dashed,orange!75!black}]
\node[block] (inputs) {Workspace and wrist images,\\instruction, state};
\node[block,below=of inputs] (embed) {Input embeddings: image tokens of\\each camera, instruction tokens,\\projected state};
\node[block,below=of embed,minimum height=22mm] (prefix) {Vision--language transformer\\mixes the inputs by attention\\$\to$ final prefix states};
\node[block,below=of prefix] (expert) {Action expert attends to the prefix\\10 Euler steps of the velocity field};
\node[block,below=of expert] (actions) {Action chunk (50 steps $\times$ 7)};
\foreach \a/\b in {inputs/embed,embed/prefix,prefix/expert,expert/actions}\draw[->] (\a) -- (\b);
\node[ctx,right=of embed] (raw) {Raw context heads, one per input: mean image tokens of the workspace or wrist camera, mean instruction token (960 values each), normalized state (8 values)};
\node[ctx,anchor=north west] (fused) at ([xshift=9mm]prefix.north east) {Fused context heads: the same four token groups in the final prefix states};
\node[tap,anchor=south west] (kvctx) at ([xshift=9mm]prefix.south east) {\KVctx{}: fused means of both cameras and the instruction, plus the 8 state values $\to$ 2,888 values};
\node[tap,right=of expert] (kvdec) {\KVdec{}: final hidden state at flow step 9, mean over 50 action tokens $\to$ 480 values};
\draw[readctx] (embed.east) -- (raw.west);
\draw[readctx] (fused.west -| prefix.east) -- (fused.west);
\draw[read] (kvctx.west -| prefix.east) -- (kvctx.west);
\draw[read] (expert.east) -- (kvdec.west);
\end{tikzpicture}
\caption{Uncertainty heads on a frozen SmolVLA policy. Dashed arrows read embeddings without changing the policy. The blue heads estimate the epistemic uncertainty of the whole observation. The orange context-specific heads estimate it for one input each, before the transformer mixes the inputs (raw) or after it (fused).}
\label{fig:architecture}
\end{figure}
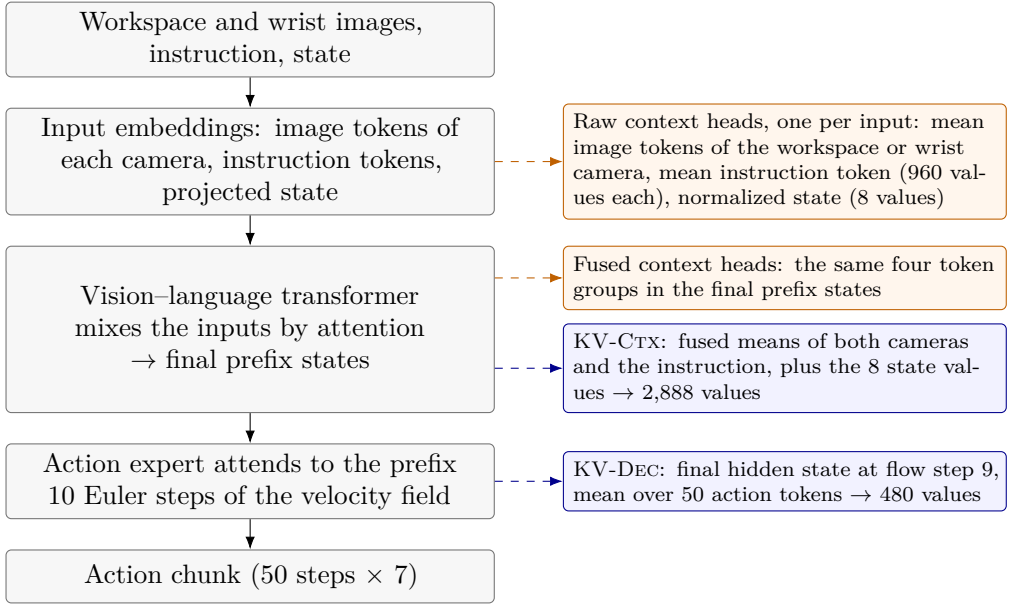

\paragraph{Policy and embeddings.}
An observation $o$ contains a workspace camera image, a wrist camera image, a language instruction, and the robot's proprioceptive state.
A condition is a starting situation from which a policy is executed.
SmolVLA~\citep{shukor2025smolvla} encodes the images, the instruction tokens and a projection of the state into a prefix with a vision--language backbone.
An action expert attends to this prefix and integrates a flow-matching velocity field over ten Euler steps, from Gaussian noise to a chunk of 50 seven-dimensional actions.
We call the policy whose uncertainty is estimated the primary.
SAVE fine-tunes it, and referral hands the conditions in which it is most uncertain to a stronger supervisor policy.
For a representation location $b$ in the policy, $\phi_b(o)$ denotes the embedding of $o$ at that location.
An uncertainty head reads such an embedding without changing the policy (Figure~\ref{fig:architecture}).

\paragraph{Support.}
The support set $\mathcal{S}$ contains observations from the demonstrations the policy is trained on, embedded by the same policy.
When demonstrations are added or the policy is retrained, the support is embedded again and the head is refitted.
No outcome of an execution, whether of the policy, of a supervisor or in evaluation, enters the fit.

\paragraph{Conditional variance.}
A head reads the embedding $\phi_b(o)$ after a preprocessing map $P_b$ fitted on the support, so its input is $z=P_b\phi_b(o)$.
Let $Z=(z_1,\ldots,z_N)$ be the preprocessed support inputs.
With a Gaussian kernel $k$ of bandwidth $\ell$ and ridge $\lambda$, the head reports the conditional variance
\begin{align}
 k(z,z')&=\exp\!\left(-\frac{\lVert z-z'\rVert^2}{2\ell^2}\right),\qquad
 u_{\KV}(z)=k(z,z)-k(z,Z)\,(K_{ZZ}+\lambda I)^{-1}\,k(Z,z),
 \label{eq:kv}
\end{align}
with $K_{ZZ}$ the kernel matrix of the support.
This is the posterior variance of a zero-mean GP with prior variance one after observing noisy values at the support inputs~\citep{rasmussen2006gpml}.
Computing $u_{\KV}$ requires no output values.
The variance $u_{\KV}(z)$ lies in $(0,1]$, is close to one far from all support inputs, and falls to $\lambda/(1+\lambda)$ at an isolated support input.
A demonstration added to the support therefore lowers the score of its own observation, and of similar observations, as soon as the head is refitted.
The score measures coverage in the chosen geometry and is computed with a Cholesky factorization of $K_{ZZ}+\lambda I$.
Exact conditioning remains affordable for the supports used here, which hold at most about 10,000 inputs.

\paragraph{Geometry.}
The preprocessing map $P_b$ standardizes each embedding coordinate over the support, optionally projects the result onto its leading principal components, and whitens it with the Ledoit--Wolf shrinkage covariance~\citep{ledoit2004covariance},
\begin{equation}
 \widehat\Sigma_\alpha=(1-\alpha)\widehat\Sigma+\alpha\,\frac{\tr\widehat\Sigma}{d}\,I,
 \qquad z=\widehat\Sigma_\alpha^{-1/2}(x-\bar x).
 \label{eq:whitening}
\end{equation}
Here $x$ is the standardized and projected embedding, $\bar x$ and $\widehat\Sigma$ are the mean and sample covariance of its $d$ coordinates over the support, and the shrinkage intensity $\alpha$ is estimated from the support.
An isotropic kernel on whitened features then uses a regularized Mahalanobis distance~\citep{lee2018mahalanobis}.

\paragraph{Two observation heads.}
Both heads read late representations of the policy, which integrate all inputs and lie close to the action output, so that distances in them should reflect the information the policy uses to choose its actions.
Appendix Table~\ref{tab:heads} summarizes the two configurations.
\KVctx{} reads the final normalized prefix states.
It averages the tokens of each camera and of the instruction, and appends the eight state coordinates, giving $3\times960+8=2{,}888$ values.
These are reduced to 512 principal components and whitened.
Its bandwidth makes the kernel equal to $0.5$ at the median distance between a support input and its $k$-th nearest support neighbour, with $k=\lceil\sqrt N\,\rceil$, and $\lambda=10^{-3}$.
\KVdec{} reads the action expert's final hidden state at the ninth of ten flow evaluations, from a fixed Gaussian initial chunk, and averages it over the 50 action tokens (480 values).
It uses full-covariance whitening, the median pairwise distance as bandwidth, and $\lambda=10^{-2}$.
\KVctx{} needs one forward pass of the prefix, and \KVdec{} additionally runs nine steps of the action expert.

\paragraph{Context-specific heads.}\label{sec:kernel-context}
The same estimator can be attached to the token groups of a single modality, such as the pooled tokens of one camera or of the instruction, at any layer.
Each such head has its own support embeddings, preprocessing and bandwidth, which enables independently attributing unfamiliarity to specific parts of the input.
We use eight such heads, all with the \KVctx{} recipe (Table~\ref{tab:heads}).
Four \emph{raw} heads read an input before the transformer mixes the inputs: the mean token embedding of the workspace camera, of the wrist camera and of the instruction (960 values each), and the eight normalized state values.
Four \emph{fused} heads read the same token groups in the final prefix states, where attention has mixed them: the two camera means and the instruction mean that \KVctx{} concatenates, and the final state at the state token.
A fused head therefore also responds to perturbations of other inputs, and the size of this response measures how strongly the policy mixes its inputs.

\subsection{Demonstration acquisition with SAVE}
\label{sec:acquisition}
SAVE~\citep{romer2026uqvla} fine-tunes a policy over several rounds.
In each round it scores the initial observation $o_{kl}$ of every candidate condition $l$ of each target task $k$, averages the scores into a task score $U_k$, and draws a few tasks with probability proportional to $U_k^{\tau}$.
For each draw, it requests the demonstration that starts from the most uncertain candidate condition of the drawn task.
The policy is then warm-started from the previous round and fine-tuned, sampling preparatory and acquired data in equal proportion.
We keep the implementation of~\citet{romer2026uqvla}, including its eligibility rule, under which previously acquired and preparatory demonstrations remain candidates, so a query can return a demonstration the policy already has.
We therefore count both query slots and genuinely new demonstrations.
The only change is the score.
Our heads use $\tau=2.5$, the temperature of SAVE/VFD, without tuning. SAVE/GU keeps the temperature $\tau=1$ and score shift of the original study.
Before each round's scoring, the support, including the acquired demonstrations, is embedded with the newly trained policy and the kernel head is refitted.

\subsection{Evaluation metrics}
\label{sec:referral}
\paragraph{Task-level calibration.}
Following~\citet{romer2026uqvla}, we correlate each task's mean uncertainty over its candidate conditions with the primary policy's success rate on that task.
We report negative Spearman ($-\rho_{\rm S}$) and Pearson ($-r_{\rm P}$) correlations, so higher is better.
As a reference for what task-level calibration rewards, we add a score that encodes only how many training demonstrations each task has, $u_k=1/(1+n_k)$, where $n_k$ counts the unique demonstrations of task $k$ in the support.
It requires task labels and says nothing about unseen tasks or perturbed scenes, so it is not a usable uncertainty estimate.

\paragraph{Supervisor referral.}
The primary and the supervisor are evaluated on the same conditions $i=1,\ldots,n$, with success rates $p_i$ and $s_i$ over repeated executions.
Referring the fraction $q$ of conditions with the highest uncertainty, $R_q$, to the supervisor yields the combined success
\begin{equation}
 A_u(q)=\frac1n\sum_{i=1}^n\left[p_i+\mathbf 1\{i\in R_q\}\,(s_i-p_i)\right].
\end{equation}
How much referral can gain depends on the supervisor's advantage and on how it is spread over the conditions, which differ between checkpoints, seeds and tasks.
To compare scores across these settings, we measure a score's gain over random referral relative to the gain of a hindsight oracle.
Random referral achieves $A_{\rm rand}(q)=(1-q)\bar p+q\bar s$ in expectation.
A hindsight oracle refers by the signed gain $g_i=s_i-p_i$, including negative gains.
The normalized area
\begin{equation}
 \operatorname{nAUC}(u)=\frac{\int_0^1[A_u(q)-A_{\rm rand}(q)]\,dq}{\int_0^1[A_{\rm oracle}(q)-A_{\rm rand}(q)]\,dq}
 \label{eq:nauc}
\end{equation}
is zero for random and one for oracle ordering. Exact score ties receive their expected value under random tie order.
\emph{Global} referral ranks all conditions together and can exploit task identity.
\emph{Within-task} referral applies the same fraction inside every task against the corresponding within-task random and oracle curves, so a score that is constant within each task scores zero.
The normalized area summarizes the whole referral curve and is our main measure.
As an intuitive example of a single operating point, we further define the automated fraction as the largest $1-q$ whose combined success stays within two percentage points of the all-supervisor success.

\paragraph{Within-task diagnostics.}
Because referral also depends on the supervisor, we report the within-task failure AUROC of the primary, averaged over tasks with mixed outcomes, using fractional failure labels $1-p_i$.
We decompose the variance of single-execution outcomes into differences between tasks, reproducible differences between conditions of a task (corrected for the sampling variance of three-execution means), and differences between executions of a condition.
As a reference for how much a single observed outcome reveals, a cross-replicate ranking orders conditions by the signed gain observed in one execution of each policy and is evaluated on the two other executions, averaged over the three choices.

\subsection{Experimental protocol}
\label{sec:protocol}
We follow the setup of~\citet{romer2026uqvla} with their released code, built on LeRobot~\citep{cadene2026lerobot}.
We repeat the study with three seeds.
For each seed, a primary and a second policy that only the ensemble scores use were trained for 30,000 updates (batch size 32) on 1,437 demonstrations: the 30 LIBERO-Spatial, -Object and -Goal tasks and LIBERO-10 tasks 0--2.
The ten LIBERO-10 tasks are the target tasks, and their 379 demonstrations form the candidate pool.
The heads' support holds the initial observation of every unique preparatory or acquired demonstration, because acquisition and referral are decided before execution.
With 1,437 to 1,512 support inputs, fitting a head takes well under a second on a GPU.
Each seed also has its own supervisor with the primary's architecture, trained on all 1,693 demonstrations of the 40 tasks for 39,993 updates, which matches the primaries' number of passes over their data.

\paragraph{Acquisition.}
For each seed we run random acquisition (named Random), which draws tasks and their candidate conditions uniformly, and SAVE/VFD, SAVE/GU, SAVE/\KVctx{} and SAVE/\KVdec{}.
Each arm uses 15 rounds of five queries and 4,000 updates per round, Random included.
We evaluate the primary at rounds 0, 5, 10 and 15 with the released evaluator of~\citet{romer2026uqvla}, using 30 rollouts per task and a 520-step horizon.
The mean success averages success over the 75 query slots, interpolating linearly between these rounds, and therefore also rewards arms that improve early.

\paragraph{Scoring on common policies.}
Calibration and referral use the primaries of the Random arm at rounds 0 and 10, so all scores see identical policies, inputs and outcomes.
Each checkpoint contributes ten tasks to the calibration.
Each checkpoint and each supervisor is executed three times, with independent seeds, from each of the 50 standard LIBERO initial states of every task, and a condition's success rate averages these executions.
Primary and supervisor start from identical initial states and receive identical initial observations, which we check by hashing the paired inputs.
The seven baselines are those of~\citet{romer2026uqvla}, computed with their code.
VFD, GU, Action-L2 and DECU measure disagreement between the two policies.
ACE (action-chunk entropy)~\citep{romer2025fiper}, VLM token entropy~\citep{malinin2021autoregressive} and VLM perplexity~\citep{fadeeva2024factcheck} use one policy and mainly reflect the spread of its own predictions, which is largely aleatoric.
We refer to them as entropy-based scores.
VFD uses the symmetric score for evaluation and, as in their code, the one-way score for acquisition.

\paragraph{Statistics and compute.}
We report means and sample standard deviations over the three seeds.
Confidence intervals are 95\% hierarchical bootstrap intervals from 2,000 draws.
Each draw resamples the 50 conditions within each task, keeping their three executions together, and then resamples the three seeds. The draws are shared across scores and checkpoints, so differences are paired.
Round-averaged values weight rounds 0 and 10 equally.
Compute times are median wall-clock times of completed single-GPU jobs (RTX 3090 or A100) in the study's job logs.

\paragraph{Context-head experiments.}
These experiments test whether the heads detect perturbations in input modalities, both in the starting observation and during execution, and whether they support referral within a task (Sections~\ref{sec:results-rollout} and~\ref{sec:results-perturbed}).
For each of three seeds we train a primary and a second policy for VFD on all 379 LIBERO-10 demonstrations, with the training setup of the main study and the same number of passes over the data (about 4.68, here 14,840 updates).
All heads are fitted post hoc on the frozen primary, once per seed, and every experiment below uses these same heads.
Their support consists of every tenth frame of its training demonstrations, because the heads also score observations during execution.
Kernel scores at the variance floor differ only by single-precision rounding, so scores of a head that lie within $10^{-6}$ of each other are treated as ties before any ranking.
Each experiment uses part of a library of 20 perturbation families, about half of them at two levels (Appendix Table~\ref{tab:perturbations}).
To test whether the uncertainty heads detect perturbed starting observations, we score the clean starting observations of half of the standard LIBERO-10 initial states of every task and their perturbed versions, without executing the policy.
To test whether the heads detect perturbations during execution, the primary runs from two initial states of every task once without and once with each perturbation, which starts at step 100. Both runs share the environment seed and policy noise.
For referral, each seed also has a supervisor trained on the same demonstrations as the primary, but with each training sample perturbed with probability 0.5 in a way that leaves the demonstrated actions correct, such as a paraphrase, colour jitter or a blank wrist image (Appendix~\ref{app:context}).
Primary and supervisor are each executed three times from ten initial states of every task, once clean and once with each of these perturbations, using the same environment seeds for both.
The 95\% confidence intervals are hierarchical bootstrap intervals from 2,000 draws.
Each draw resamples the initial states within each task and then the three seeds.
An initial state keeps all its perturbed versions, because they start from the same scene.
For perturbations during execution, which use only two initial states per task, each draw resamples the pairs of task and initial state instead.
The intervals are not adjusted for multiple comparisons, so comparisons for individual perturbations are interpreted as exploratory.

\section{Results}
\label{sec:results}

We first compare our uncertainty scoring heads with ensemble scores from the literature in our reproduction of the SAVE study, in task-level calibration (Q1, Section~\ref{sec:results-calibration}), demonstration acquisition and its cost (Q2 and Q3, Section~\ref{sec:results-acquisition}) and supervisor referral (Q1, Section~\ref{sec:results-referral}).
We then examine what these evaluations actually measure, in particular how well they test the ranking of situations within a task (Q4, Section~\ref{sec:results-benchmark}).
To evaluate this ranking directly, we create differences within a task by controlled input perturbations.
We test whether the uncertainty scores detect these perturbations, whether uncertainty heads on single input modalities show which input was perturbed, and whether referring the most uncertain conditions of each task to the supervisor improves success over random referral (Q5, Sections~\ref{sec:results-rollout} and~\ref{sec:results-perturbed}).

\subsection{Task-level calibration}
\label{sec:results-calibration}

\begin{table}[t]
\centering\small
\caption{Task-level calibration: negative Spearman and Pearson correlation between task-mean uncertainty and task success. All scores are evaluated on the same policies, before acquisition and after ten rounds of random acquisition (Section~\ref{sec:protocol}), averaged over these two checkpoints (mean $\pm$ SD over three seeds, best learned score in bold). Policies: number of trained policies a score needs. Type E marks disagreement between two policies, A an entropy-based score of one policy (mainly aleatoric) and K a kernel variance. Reported: values of \citet{romer2026uqvla}, averaged over 15 rounds of a separate 2,000-update run, so the protocols differ. $^\dagger$Reference that uses task labels, not a usable score.}
\label{tab:calibration}
\begin{tabular}{@{}lcccccc@{}}
\toprule
& & & \multicolumn{2}{c}{This study} & \multicolumn{2}{c}{Reported}\\
\cmidrule(lr){4-5}\cmidrule(l){6-7}
Score & Policies & Type & $-\rho_{\rm S}$ & $-r_{\rm P}$ & $-\rho_{\rm S}$ & $-r_{\rm P}$\\
\midrule
VFD & 2 & E & $0.75\pm0.12$ & $0.64\pm0.17$ & $0.71$ & $0.71$\\
GU & 2 & E & $0.66\pm0.19$ & $0.68\pm0.20$ & $0.62$ & $0.65$\\
Action-L2 & 2 & E & $0.56\pm0.07$ & $0.50\pm0.15$ & $0.50$ & $0.48$\\
ACE & 1 & A & $0.43\pm0.16$ & $0.44\pm0.16$ & $0.31$ & $0.36$\\
DECU & 2 & E & $0.37\pm0.29$ & $0.40\pm0.08$ & $0.31$ & $0.23$\\
VLM entropy & 1 & A & $-0.09\pm0.19$ & $-0.03\pm0.10$ & $0.10$ & $0.23$\\
VLM perplexity & 1 & A & $-0.14\pm0.25$ & $0.00\pm0.27$ & $-0.04$ & $0.02$\\
\midrule
\KVctx{} & 1 & K & {\boldmath$0.77\pm0.08$} & {\boldmath$0.82\pm0.09$} & -- & --\\
\KVdec{} & 1 & K & $0.75\pm0.17$ & $0.80\pm0.14$ & -- & --\\
\midrule
Demonstration count$^\dagger$ & -- & -- & $0.82\pm0.18$ & $0.83\pm0.17$ & -- & --\\
\bottomrule
\end{tabular}

\end{table}

Table~\ref{tab:calibration} first verifies our reproduction of the baselines.
To save computation, we score two checkpoints of our random-acquisition run instead of all fifteen rounds of a separate calibration run as~\citet{romer2026uqvla} (Section~\ref{sec:protocol}).
The seven baselines nevertheless rank in the same order as reported, where ACE and DECU were tied, and most values lie within about 0.1 of theirs.
VFD reaches $-\rho_{\rm S}=0.75\pm0.12$, so the reported $0.71$, also from three seeds, lies well within the spread over our seeds.

Both our single-policy heads are on par with VFD.
\KVctx{} reaches $-\rho_{\rm S}=0.77\pm0.08$ and $-r_{\rm P}=0.82\pm0.09$.
The paired difference to VFD is $+0.02$ $[-0.02,0.09]$ in Spearman and $+0.18$ $[0.11,0.25]$ in Pearson correlation.
\KVdec{} behaves similarly (Appendix Table~\ref{tab:paired}).
The Pearson advantage is largest at round 0, where \KVctx{} reaches $-r_{\rm P}=0.94$ and VFD $0.75$ (Appendix Table~\ref{tab:per-round}).
In the SAVE LIBERO-10 protocol, before acquisition, three of the ten target tasks are already part of the training data, so the head assigns their candidate conditions low variance.
The seven tasks without demonstrations score much higher, and this clean split between the two groups matches the success pattern almost linearly.

The last row of Table~\ref{tab:calibration} does not estimate uncertainty but counts how many demonstrations of each task the policy has been trained on.
This reference correlates with task success at least as strongly as every learned score ($-\rho_{\rm S}=0.82$, $-r_{\rm P}=0.83$).
We return to this observation in Section~\ref{sec:results-benchmark}.

\subsection{Demonstration acquisition and cost}
\label{sec:results-acquisition}

\begin{figure}[t]
\centering
\includegraphics[width=0.62\linewidth]{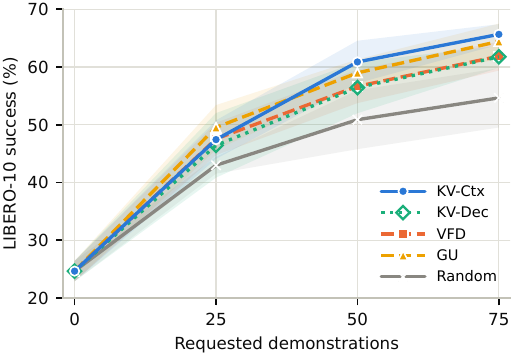}
\caption{SAVE learning curves on LIBERO-10. Success of the fine-tuned policy (30 evaluation rollouts per task) as demonstrations are requested, mean and $\pm1$ SD band over three independent runs.}
\label{fig:acquisition}
\end{figure}

\begin{table}[t]
\centering\small
\setlength{\tabcolsep}{2.5pt}
\caption{Policy success and cost during demonstration acquisition (mean $\pm$ SD over three seeds, best in bold). All arms use five queries and 4,000 updates per round. Success: after 5, 10 and 15 rounds, and the normalized area under the linearly interpolated learning curve (Mean). New: unique previously unseen demonstrations acquired after 75 queries. Overhead: median GPU-minutes per round for scoring the candidate pool. For ensemble scores, it also includes fine-tuning the second policy for 4,000 updates per round, but not the one-time pre-training of that policy (30,000 updates). The training of the acquiring policy itself (about 37 minutes per round) is common to all arms and excluded.}
\label{tab:acquisition}
\begin{tabular}{@{}lccccccc@{}}
\toprule
& & \multicolumn{4}{c}{Success (\%)} & & Overhead\\
\cmidrule(lr){3-6}
Acquisition & Policies & 5 & 10 & 15 & Mean & New & (min/round)\\
\midrule
Random & 1 & $43.0\pm1.5$ & $50.9\pm5.1$ & $54.7\pm5.1$ & $44.5\pm2.7$ & $46.3\pm6.7$ & --\\
VFD & 2 & $47.6\pm2.8$ & $56.7\pm2.9$ & $61.9\pm2.5$ & $49.2\pm0.6$ & $69.3\pm4.5$ & $43.8$\\
GU & 2 & {\boldmath$49.6\pm3.9$} & $59.0\pm2.3$ & $64.4\pm3.1$ & $51.0\pm1.0$ & $62.7\pm2.5$ & $44.6$\\
\midrule
\KVctx{} & 1 & $47.4\pm3.3$ & {\boldmath$60.9\pm3.7$} & {\boldmath$65.7\pm1.8$} & {\boldmath$51.2\pm1.6$} & $75.0\pm0.0$ & $1.0$\\
\KVdec{} & 1 & $46.4\pm5.6$ & $56.4\pm4.4$ & $61.8\pm1.9$ & $48.7\pm3.2$ & $75.0\pm0.0$ & $12.0$\\
\bottomrule
\end{tabular}

\end{table}

Figure~\ref{fig:acquisition} and Table~\ref{tab:acquisition} compare the acquisition strategies.
Every uncertainty-guided arm ends above Random ($54.7\%$).
SAVE with \KVctx{} ends at $65.7\pm1.8\%$, with a mean success of $51.2\%$ over the run.
This is level with SAVE/GU ($64.4\pm3.1\%$, mean $51.0\%$) and above SAVE/VFD ($61.9\pm2.5\%$, mean $49.2\%$).
These results show similar observed acquisition performance, although three seeds do not establish equivalence or a consistent performance advantage among these three arms.
\KVdec{} ($61.8\pm1.9\%$) also has a similar final success rate to VFD.
Random and GU reproduce the reported final success rates of $54.6\%$ and $64.0\%$, whereas our VFD arm remains below the reported $67.1\%$~\citep{romer2026uqvla}.

The Gaussian kernel heads spend the query budget differently.
Because an acquired observation drops to the variance floor set by the ridge regularization at the next refit, in our experiments, SAVE never requested a demonstration twice or queried the preparatory demonstrations, so all 75 queries returned unique new demonstrations, compared with an average $69.3$ for VFD, $62.7$ for GU and $46.3$ for Random.

The heads remove most of the computational overhead of uncertainty-guided acquisition.
Per round, \KVctx{} needs about one GPU-minute to embed the support and candidates, refit and score, whereas VFD needs 9.0 minutes of ensemble scoring plus 34.8 minutes to fine-tune the second policy, which must also be pre-trained once for 30,000 updates (about 4 GPU-hours).
Over 15 rounds this amounts to about 15 GPU-minutes versus 15 GPU-hours (including the one-time pre-training of the second policy).
At deployment, once all policies are trained, scoring the 500 evaluation conditions of one checkpoint takes 0.8 GPU-minutes with \KVctx{} (2.2\,GiB of GPU memory), 8.9 with \KVdec{}, 13.8 with GU and 38.8 with VFD.

\subsection{Supervisor referral}
\label{sec:results-referral}

\begin{figure}[t]
\centering
\includegraphics[width=\linewidth]{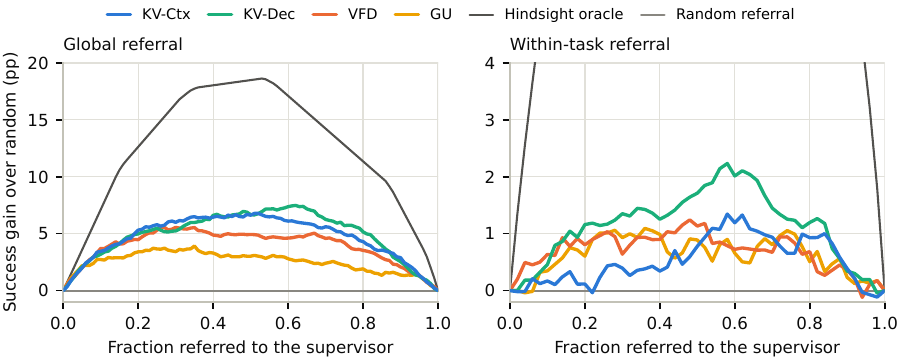}
\caption{Referral to the supervisor at round 10: combined success minus random referral, as a function of the fraction of conditions referred (mean over three seeds). The hindsight oracle refers by the observed signed gain. In the within-task panel it peaks at 14.1 percentage points, outside the plotted range.}
\label{fig:referral}
\end{figure}

\begin{table}[t]
\centering\small
\setlength{\tabcolsep}{5pt}
\caption{Referral from the policies of Table~\ref{tab:calibration} to a stronger supervisor policy (mean $\pm$ SD over three seeds, best learned score in bold). Global nAUC and automated fraction average acquisition rounds 0 and 10. Within-task nAUC refers to round 10. Chance level is 0. $^\dagger$Conditions ranked by the demonstration count of their task, a reference that uses task labels and is constant within a task (Table~\ref{tab:calibration}). $^\ddagger$Conditions ranked by the supervisor's gain observed in one of the three executions of each condition and evaluated on the other two. It shows how much a single observed outcome reveals (Section~\ref{sec:referral}).}
\label{tab:referral}
\begin{tabular}{@{}lccc@{}}
\toprule
& Global & Within-task & Automated\\
Score & nAUC & nAUC (r10) & fraction\\
\midrule
VFD & $0.48\pm0.06$ & $0.07\pm0.05$ & $0.21\pm0.03$\\
GU & $0.42\pm0.05$ & $0.06\pm0.03$ & $0.19\pm0.04$\\
Action-L2 & $0.35\pm0.01$ & $0.06\pm0.04$ & $0.13\pm0.03$\\
ACE & $0.29\pm0.06$ & $0.06\pm0.10$ & $0.13\pm0.03$\\
DECU & $0.20\pm0.11$ & $0.03\pm0.01$ & $0.11\pm0.01$\\
VLM entropy & $-0.06\pm0.09$ & $0.02\pm0.03$ & $0.05\pm0.02$\\
VLM perplexity & $-0.13\pm0.08$ & $-0.10\pm0.01$ & $0.04\pm0.01$\\
\midrule
\KVctx{} & $0.50\pm0.04$ & $0.05\pm0.12$ & $0.26\pm0.03$\\
\KVdec{} & {\boldmath$0.51\pm0.07$} & {\boldmath$0.11\pm0.03$} & {\boldmath$0.30\pm0.05$}\\
\midrule
Demonstration count$^\dagger$ & $0.52\pm0.07$ & -- & $0.32\pm0.03$\\
Cross-replicate reference$^\ddagger$ & -- & $0.27\pm0.04$ & --\\
\bottomrule
\end{tabular}

\end{table}

On the 500 evaluation conditions (the 50 standard initial states of each of the ten target tasks), the supervisors succeed in 78--81\% of the executions, the primaries in 24\% at round 0 and in 47--53\% at round 10.
Globally, the kernel heads match or exceed every ensemble score in referral to the supervisor (Table~\ref{tab:referral}, Figure~\ref{fig:referral}).
\KVctx{} reaches a normalized area of $0.50\pm0.04$, \KVdec{} $0.51\pm0.07$, VFD $0.48\pm0.06$ and GU $0.42\pm0.05$.
In paired comparisons, the 95\% intervals of both heads' differences from VFD include zero, whereas against GU they lie above zero, which supports better referral by both heads than by GU (Appendix Table~\ref{tab:paired}).
For the automated fraction, the intervals likewise support both heads over GU and \KVdec{} over VFD.
The entropy-based scores (ACE, VLM entropy, VLM perplexity) refer poorly.

Referral becomes harder after acquisition: VFD, GU and both kernel heads reach areas of $0.63$--$0.65$ at round 0, but $0.38$ (\KVctx{}), $0.39$ (\KVdec{}), $0.31$ (VFD) and $0.20$ (GU) at round 10 (Appendix Table~\ref{tab:per-round}).
Within tasks, no score refers well: the best within-task nAUC at round 10 is $0.11\pm0.03$ for \KVdec{}, and its interval $[0.04,0.20]$ is the only one in Table~\ref{tab:referral} that lies above zero.
The next section examines why.

\subsection{Task coverage versus within-task ranking}
\label{sec:results-benchmark}
\begin{figure}[t]
\centering
\includegraphics[width=\linewidth]{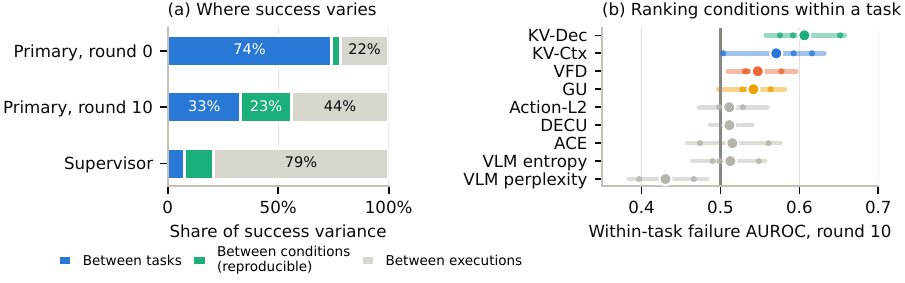}
\caption{(a) Share of the variance of single-execution success that lies between tasks, between conditions of a task (corrected for the sampling noise of three-execution means), and between executions of a condition, averaged over three seeds. (b) Within-task failure AUROC at round 10: seed values (small dots), mean (large dot) and 95\% hierarchical bootstrap interval (bar). Chance is 0.5.}
\label{fig:benchmark}
\end{figure}

Task-level calibration mainly reflects demonstration coverage, as indicated by the demonstration count reference in Table~\ref{tab:calibration}.
Ranking conditions by this count also matches or exceeds every uncertainty score in global referral (nAUC $0.52\pm0.07$, automated fraction $0.32\pm0.03$, Table~\ref{tab:referral}).
At round 0, the primary almost always fails the seven tasks without demonstrations but succeeds in 70--88\% of executions on the three preparatory tasks.
Here, demonstration counts alone reach $-\rho_{\rm S}=0.94$ and $-r_{\rm P}=0.997$.
Thus, high task-level correlations need not imply useful within-task rankings.

Task-level differences explain 71--76\% of outcome variance at round 0, versus only 3--7\% of reproducible within-task differences (Figure~\ref{fig:benchmark}a).
After ten rounds of random acquisition, these shares become 27--37\% between-task and 20--25\% within-task, while the remaining 44\% reflects variation between executions of the same condition.
Within-task failure AUROC at round 10 reaches $0.61$ $[0.56,0.66]$ for \KVdec{}, $0.57$ $[0.50,0.63]$ for \KVctx{}, $0.55$ $[0.51,0.60]$ for VFD and $0.54$ $[0.50,0.58]$ for GU.
Entropy-based scores remain near or below chance (Figure~\ref{fig:benchmark}b).
Execution variability also limits within-task referral.
Ranking conditions by the supervisor's gain in one execution and evaluating on the other two yields nAUC $0.27\pm0.04$ (Table~\ref{tab:referral}). Relative to this reference, \KVdec{} achieves $0.11$ (42\%), VFD $0.07$ (26\%), GU $0.06$ (23\%) and \KVctx{} $0.05$ (19\%).

Kernel and ensemble scores agree on coverage between tasks but not within tasks.
\KVctx{}--VFD rank correlation is $0.81$ globally at round 0 and $0.59$ at round 10, but only $0.25$--$0.27$ within tasks. Within-task agreement is stronger within families: $0.74$--$0.78$ for VFD--GU and $0.67$--$0.68$ between kernel heads.
Within a task, the two families therefore single out different conditions, and the noisy outcomes cannot show which choice is more useful.

Thus, LIBERO's standard initial conditions, scored before execution as in SAVE, test between-task coverage well but within-task coverage only weakly, motivating the targeted evaluations below.

\subsection{Context-specific heads under controlled perturbations}
\label{sec:results-rollout}

\begin{table}[t]
\centering\footnotesize
\setlength{\tabcolsep}{2.5pt}
\caption{Detection of perturbed starting observations: AUROC of perturbed against clean starting observations within a task for the context and whole-policy heads of Figure~\ref{fig:architecture} and VFD, mean over three seeds. Each of the LIBERO-10 tasks contributes 25 initial states, of which the more expensive VFD was evaluated on only 10. Chance level is 0.5.}
\label{tab:context-start}
\begin{tabularx}{\linewidth}{@{}r@{\hspace{4pt}}>{\raggedright\arraybackslash}X*{4}{c}!{\hspace{2pt}}*{4}{c}!{\hspace{2pt}}cc!{\hspace{2pt}}c@{}}
\toprule
& & \multicolumn{4}{c}{Raw heads} & \multicolumn{4}{c}{Fused heads} & \multicolumn{3}{c}{Whole-policy}\\
\cmidrule(lr){3-6}\cmidrule(lr){7-10}\cmidrule(l){11-13}
& Perturbation & \rotatebox{90}{Workspace} & \rotatebox{90}{Wrist} & \rotatebox{90}{Instruction} & \rotatebox{90}{State} & \rotatebox{90}{Workspace} & \rotatebox{90}{Wrist} & \rotatebox{90}{Instruction} & \rotatebox{90}{State} & \rotatebox{90}{\KVctx{}} & \rotatebox{90}{\KVdec{}} & \rotatebox{90}{VFD}\\
\midrule
\multicolumn{13}{@{}l}{\textit{Instruction}}\\
S1 & Strongly paraphrased & \cellcolor[HTML]{F0EFEC}0.50 & \cellcolor[HTML]{F0EFEC}0.50 & \cellcolor[HTML]{86B6F0}1.00 & \cellcolor[HTML]{F0EFEC}0.50 & \cellcolor[HTML]{EBEDED}0.53 & \cellcolor[HTML]{E6EBEE}0.55 & \cellcolor[HTML]{88B7F0}0.99 & \cellcolor[HTML]{8CB9F1}0.98 & \cellcolor[HTML]{88B7F0}0.99 & \cellcolor[HTML]{86B6F0}1.00 & \cellcolor[HTML]{91BCF1}0.95\\
S2 & All object words replaced by synonyms & \cellcolor[HTML]{F0EFEC}0.50 & \cellcolor[HTML]{F0EFEC}0.50 & \cellcolor[HTML]{86B6F0}1.00 & \cellcolor[HTML]{F0EFEC}0.50 & \cellcolor[HTML]{EEEEEC}0.51 & \cellcolor[HTML]{EDEEED}0.52 & \cellcolor[HTML]{94BEF2}0.94 & \cellcolor[HTML]{91BCF1}0.95 & \cellcolor[HTML]{96C0F2}0.93 & \cellcolor[HTML]{87B7F0}0.99 & \cellcolor[HTML]{99C1F2}0.92\\
S3 & One typo in a non-object word & \cellcolor[HTML]{F0EFEC}0.50 & \cellcolor[HTML]{F0EFEC}0.50 & \cellcolor[HTML]{86B6F0}1.00 & \cellcolor[HTML]{F0EFEC}0.50 & \cellcolor[HTML]{F0EFEC}0.50 & \cellcolor[HTML]{EEEEEC}0.51 & \cellcolor[HTML]{A3C7F3}0.88 & \cellcolor[HTML]{A9CBF4}0.85 & \cellcolor[HTML]{A8CAF4}0.86 & \cellcolor[HTML]{8DBAF1}0.97 & \cellcolor[HTML]{B2D0F5}0.82\\
S4 & Object nouns replaced by pronouns & \cellcolor[HTML]{F0EFEC}0.50 & \cellcolor[HTML]{F0EFEC}0.50 & \cellcolor[HTML]{86B6F0}1.00 & \cellcolor[HTML]{F0EFEC}0.50 & \cellcolor[HTML]{D9E4F2}0.63 & \cellcolor[HTML]{D6E3F2}0.65 & \cellcolor[HTML]{86B6F0}1.00 & \cellcolor[HTML]{86B6F0}1.00 & \cellcolor[HTML]{86B6F0}1.00 & \cellcolor[HTML]{86B6F0}1.00 & \cellcolor[HTML]{8DBAF1}0.97\\
S5 & Translated into German & \cellcolor[HTML]{F0EFEC}0.50 & \cellcolor[HTML]{F0EFEC}0.50 & \cellcolor[HTML]{86B6F0}1.00 & \cellcolor[HTML]{F0EFEC}0.50 & \cellcolor[HTML]{C1D9F7}0.76 & \cellcolor[HTML]{ADCDF5}0.84 & \cellcolor[HTML]{86B6F0}1.00 & \cellcolor[HTML]{88B7F0}0.99 & \cellcolor[HTML]{86B6F0}1.00 & \cellcolor[HTML]{86B6F0}1.00 & \cellcolor[HTML]{87B7F0}0.99\\
\addlinespace[2pt]
\multicolumn{13}{@{}l}{\textit{Camera images}}\\
S6 & Light at 40\,\% intensity & \cellcolor[HTML]{A7CAF4}0.86 & \cellcolor[HTML]{90BCF1}0.96 & \cellcolor[HTML]{F0EFEC}0.50 & \cellcolor[HTML]{F0EFEC}0.50 & \cellcolor[HTML]{9CC3F3}0.91 & \cellcolor[HTML]{90BCF1}0.96 & \cellcolor[HTML]{DEE7F0}0.60 & \cellcolor[HTML]{D6E3F2}0.64 & \cellcolor[HTML]{B6D2F5}0.80 & \cellcolor[HTML]{ACCDF4}0.84 & \cellcolor[HTML]{BDD6F6}0.77\\
S7 & Gaussian noise on both images ($\sigma=0.08$) & \cellcolor[HTML]{D2E1F3}0.67 & \cellcolor[HTML]{B1CFF5}0.82 & \cellcolor[HTML]{F0EFEC}0.50 & \cellcolor[HTML]{F0EFEC}0.50 & \cellcolor[HTML]{D0E0F4}0.68 & \cellcolor[HTML]{AFCEF5}0.83 & \cellcolor[HTML]{EEEEED}0.51 & \cellcolor[HTML]{EFEFEC}0.51 & \cellcolor[HTML]{E2E9EF}0.58 & \cellcolor[HTML]{E0E8F0}0.59 & \cellcolor[HTML]{E0E8F0}0.59\\
S8 & Workspace camera moved by 4\,cm and 6$^\circ$ & \cellcolor[HTML]{9BC2F2}0.91 & \cellcolor[HTML]{F0EEEA}0.49 & \cellcolor[HTML]{F0EFEC}0.50 & \cellcolor[HTML]{F0EFEC}0.50 & \cellcolor[HTML]{91BCF1}0.95 & \cellcolor[HTML]{E9ECEE}0.54 & \cellcolor[HTML]{D2E1F3}0.67 & \cellcolor[HTML]{CCDEF5}0.70 & \cellcolor[HTML]{ADCDF5}0.84 & \cellcolor[HTML]{A0C5F3}0.89 & \cellcolor[HTML]{B9D4F6}0.79\\
S9 & Target 75\,\% occluded in the workspace image & \cellcolor[HTML]{CEDFF4}0.69 & \cellcolor[HTML]{F0EFEC}0.50 & \cellcolor[HTML]{F0EFEC}0.50 & \cellcolor[HTML]{F0EFEC}0.50 & \cellcolor[HTML]{CADDF5}0.71 & \cellcolor[HTML]{EEEEED}0.51 & \cellcolor[HTML]{E0E8F0}0.59 & \cellcolor[HTML]{E6EBEE}0.55 & \cellcolor[HTML]{CEDFF4}0.69 & \cellcolor[HTML]{CEDFF4}0.69 & \cellcolor[HTML]{DDE6F1}0.60\\
S10 & Table texture replaced & \cellcolor[HTML]{C7DCF6}0.73 & \cellcolor[HTML]{A1C6F3}0.89 & \cellcolor[HTML]{F0EFEC}0.50 & \cellcolor[HTML]{F0EFEC}0.50 & \cellcolor[HTML]{C3DAF7}0.75 & \cellcolor[HTML]{A2C6F3}0.89 & \cellcolor[HTML]{DEE7F0}0.60 & \cellcolor[HTML]{D9E4F2}0.63 & \cellcolor[HTML]{C7DCF6}0.73 & \cellcolor[HTML]{C1D9F7}0.76 & \cellcolor[HTML]{C7DCF6}0.73\\
S11 & Wrist image blacked out & \cellcolor[HTML]{F0EFEC}0.50 & \cellcolor[HTML]{86B6F0}1.00 & \cellcolor[HTML]{F0EFEC}0.50 & \cellcolor[HTML]{F0EFEC}0.50 & \cellcolor[HTML]{D3E1F3}0.66 & \cellcolor[HTML]{86B6F0}1.00 & \cellcolor[HTML]{95BFF2}0.94 & \cellcolor[HTML]{8BB9F1}0.98 & \cellcolor[HTML]{86B6F0}1.00 & \cellcolor[HTML]{86B6F0}1.00 & \cellcolor[HTML]{86B6F0}1.00\\
\addlinespace[2pt]
\multicolumn{13}{@{}l}{\textit{Scene}}\\
S12 & Target displaced by 10\,cm & \cellcolor[HTML]{D1E0F4}0.67 & \cellcolor[HTML]{98C1F2}0.92 & \cellcolor[HTML]{F0EFEC}0.50 & \cellcolor[HTML]{F0EFEC}0.50 & \cellcolor[HTML]{CCDEF5}0.70 & \cellcolor[HTML]{99C1F2}0.92 & \cellcolor[HTML]{D3E2F3}0.66 & \cellcolor[HTML]{D1E0F4}0.67 & \cellcolor[HTML]{C0D8F7}0.76 & \cellcolor[HTML]{B0CFF5}0.82 & \cellcolor[HTML]{CDDFF5}0.69\\
S13 & Another object moved next to the target & \cellcolor[HTML]{C5DBF7}0.74 & \cellcolor[HTML]{97C0F2}0.93 & \cellcolor[HTML]{F0EFEC}0.50 & \cellcolor[HTML]{F0EFEC}0.50 & \cellcolor[HTML]{C7DCF6}0.73 & \cellcolor[HTML]{9CC3F3}0.91 & \cellcolor[HTML]{D9E4F2}0.63 & \cellcolor[HTML]{DAE5F1}0.62 & \cellcolor[HTML]{C4DAF7}0.75 & \cellcolor[HTML]{B6D2F5}0.80 & \cellcolor[HTML]{D6E3F2}0.65\\
S14 & Target rotated by 90$^\circ$ & \cellcolor[HTML]{E1E8F0}0.59 & \cellcolor[HTML]{CADDF5}0.71 & \cellcolor[HTML]{F0EFEC}0.50 & \cellcolor[HTML]{F0EFEC}0.50 & \cellcolor[HTML]{DDE6F1}0.60 & \cellcolor[HTML]{CCDEF5}0.70 & \cellcolor[HTML]{E1E8F0}0.58 & \cellcolor[HTML]{E5EAEF}0.56 & \cellcolor[HTML]{D9E4F2}0.63 & \cellcolor[HTML]{D2E1F3}0.66 & \cellcolor[HTML]{E0E8F0}0.59\\
\addlinespace[2pt]
\multicolumn{13}{@{}l}{\textit{Robot}}\\
S15 & Arm joints offset by 0.1\,rad & \cellcolor[HTML]{B9D4F6}0.79 & \cellcolor[HTML]{89B8F0}0.99 & \cellcolor[HTML]{F0EFEC}0.50 & \cellcolor[HTML]{86B6F0}1.00 & \cellcolor[HTML]{AFCEF5}0.83 & \cellcolor[HTML]{8CB9F1}0.98 & \cellcolor[HTML]{CFE0F4}0.68 & \cellcolor[HTML]{95BFF2}0.94 & \cellcolor[HTML]{A5C8F4}0.87 & \cellcolor[HTML]{8CB9F1}0.97 & \cellcolor[HTML]{95BFF2}0.94\\
S16 & Reported end-effector position offset by 3\,cm & \cellcolor[HTML]{F0EFEC}0.50 & \cellcolor[HTML]{F0EFEC}0.50 & \cellcolor[HTML]{F0EFEC}0.50 & \cellcolor[HTML]{86B6F0}1.00 & \cellcolor[HTML]{F0EFEC}0.50 & \cellcolor[HTML]{F0EFEC}0.50 & \cellcolor[HTML]{F0EFEC}0.50 & \cellcolor[HTML]{CDDFF5}0.69 & \cellcolor[HTML]{F0EFEC}0.50 & \cellcolor[HTML]{DAE5F1}0.62 & \cellcolor[HTML]{DAE5F1}0.62\\
\addlinespace[2pt]
\multicolumn{13}{@{}l}{\textit{Instruction and scene}}\\
S17 & Other task's instruction, its objects absent & \cellcolor[HTML]{F0EFEC}0.50 & \cellcolor[HTML]{F0EFEC}0.50 & \cellcolor[HTML]{F0EFEC}0.50 & \cellcolor[HTML]{F0EFEC}0.50 & \cellcolor[HTML]{E9ECEE}0.54 & \cellcolor[HTML]{E6EAEF}0.56 & \cellcolor[HTML]{B4D1F5}0.81 & \cellcolor[HTML]{89B8F0}0.99 & \cellcolor[HTML]{95BFF2}0.93 & \cellcolor[HTML]{86B6F0}1.00 & \cellcolor[HTML]{8CB9F1}0.97\\
 & from the scene & & & & & & & & & & &\\
\midrule
& Mean over perturbations & & & & & & & & & \cellcolor[HTML]{B3D1F5}0.81 & \cellcolor[HTML]{A7CAF4}0.86 & \cellcolor[HTML]{B7D3F6}0.80\\
\bottomrule
\end{tabularx}

\end{table}

\begin{table}[t]
\centering\footnotesize
\setlength{\tabcolsep}{2.5pt}
\caption{Detection of perturbations during execution: AUROC of the 50 steps after a perturbation against the same steps of an unperturbed run for the context and whole-policy heads of Figure~\ref{fig:architecture} and VFD with the same initial state and seed, for two initial states per task and averaged over three seeds. Chance level is 0.5.}
\label{tab:context-execution}
\begin{tabularx}{\linewidth}{@{}r@{\hspace{4pt}}>{\raggedright\arraybackslash}X*{4}{c}!{\hspace{2pt}}*{4}{c}!{\hspace{2pt}}cc!{\hspace{2pt}}c@{}}
\toprule
& & \multicolumn{4}{c}{Raw heads} & \multicolumn{4}{c}{Fused heads} & \multicolumn{3}{c}{Whole-policy}\\
\cmidrule(lr){3-6}\cmidrule(lr){7-10}\cmidrule(l){11-13}
& Perturbation & \rotatebox{90}{Workspace} & \rotatebox{90}{Wrist} & \rotatebox{90}{Instruction} & \rotatebox{90}{State} & \rotatebox{90}{Workspace} & \rotatebox{90}{Wrist} & \rotatebox{90}{Instruction} & \rotatebox{90}{State} & \rotatebox{90}{\KVctx{}} & \rotatebox{90}{\KVdec{}} & \rotatebox{90}{VFD}\\
\midrule
\multicolumn{13}{@{}l}{\textit{Instruction}}\\
E1 & Strongly paraphrased & \cellcolor[HTML]{EBEDED}0.52 & \cellcolor[HTML]{E6EAEF}0.56 & \cellcolor[HTML]{86B6F0}1.00 & \cellcolor[HTML]{E5EAEF}0.56 & \cellcolor[HTML]{E8EBEE}0.54 & \cellcolor[HTML]{ECEDED}0.52 & \cellcolor[HTML]{8AB8F0}0.98 & \cellcolor[HTML]{BFD8F7}0.76 & \cellcolor[HTML]{8DBAF1}0.97 & \cellcolor[HTML]{92BDF1}0.95 & \cellcolor[HTML]{9FC5F3}0.90\\
E2 & Switched to another task's instruction & \cellcolor[HTML]{DFE7F0}0.59 & \cellcolor[HTML]{D9E4F2}0.63 & \cellcolor[HTML]{F0EFEC}0.50 & \cellcolor[HTML]{CFE0F4}0.68 & \cellcolor[HTML]{DDE6F1}0.61 & \cellcolor[HTML]{DDE6F1}0.60 & \cellcolor[HTML]{B9D4F6}0.79 & \cellcolor[HTML]{BED7F6}0.77 & \cellcolor[HTML]{A3C7F3}0.88 & \cellcolor[HTML]{95BFF2}0.93 & \cellcolor[HTML]{9AC2F2}0.92\\
E3 & Negated (``do not \ldots'') & \cellcolor[HTML]{F0ECE8}0.48 & \cellcolor[HTML]{F0ECE8}0.48 & \cellcolor[HTML]{86B6F0}1.00 & \cellcolor[HTML]{F0EEEA}0.49 & \cellcolor[HTML]{F1E9E4}0.46 & \cellcolor[HTML]{F1E8E4}0.46 & \cellcolor[HTML]{9DC4F3}0.90 & \cellcolor[HTML]{DFE7F0}0.59 & \cellcolor[HTML]{AFCEF5}0.83 & \cellcolor[HTML]{BCD6F6}0.78 & \cellcolor[HTML]{D9E4F2}0.63\\
\addlinespace[2pt]
\multicolumn{13}{@{}l}{\textit{Camera images}}\\
E4 & Light at 40\,\% intensity & \cellcolor[HTML]{B7D3F6}0.80 & \cellcolor[HTML]{BDD7F6}0.77 & \cellcolor[HTML]{F0EFEC}0.50 & \cellcolor[HTML]{F1E9E4}0.46 & \cellcolor[HTML]{AACBF4}0.85 & \cellcolor[HTML]{BCD6F6}0.78 & \cellcolor[HTML]{E2E8F0}0.58 & \cellcolor[HTML]{EEEEEC}0.51 & \cellcolor[HTML]{C2DAF7}0.75 & \cellcolor[HTML]{C8DCF6}0.72 & \cellcolor[HTML]{D8E4F2}0.64\\
E5 & Workspace camera bumped by 2\,cm and 3$^\circ$ & \cellcolor[HTML]{D3E2F3}0.66 & \cellcolor[HTML]{F1E9E4}0.46 & \cellcolor[HTML]{F0EFEC}0.50 & \cellcolor[HTML]{F1E6E2}0.44 & \cellcolor[HTML]{C5DBF6}0.74 & \cellcolor[HTML]{F1E6E2}0.45 & \cellcolor[HTML]{ECEDED}0.52 & \cellcolor[HTML]{F1EAE6}0.47 & \cellcolor[HTML]{E7EBEE}0.55 & \cellcolor[HTML]{E9ECEE}0.54 & \cellcolor[HTML]{E2E8F0}0.58\\
E6 & Target 75\,\% occluded for 60 steps & \cellcolor[HTML]{D3E1F3}0.66 & \cellcolor[HTML]{F0EDE9}0.49 & \cellcolor[HTML]{F0EFEC}0.50 & \cellcolor[HTML]{F1EAE6}0.47 & \cellcolor[HTML]{C5DBF6}0.74 & \cellcolor[HTML]{F1E9E5}0.46 & \cellcolor[HTML]{D9E4F2}0.63 & \cellcolor[HTML]{EFEFEC}0.50 & \cellcolor[HTML]{CFE0F4}0.68 & \cellcolor[HTML]{D6E3F2}0.64 & \cellcolor[HTML]{DCE6F1}0.61\\
E7 & Wrist image frozen for 60 steps & \cellcolor[HTML]{E8EBEE}0.55 & \cellcolor[HTML]{F0ECE9}0.48 & \cellcolor[HTML]{F0EFEC}0.50 & \cellcolor[HTML]{E7EBEE}0.55 & \cellcolor[HTML]{E4EAEF}0.57 & \cellcolor[HTML]{F1E9E4}0.46 & \cellcolor[HTML]{E2E9EF}0.58 & \cellcolor[HTML]{E2E9EF}0.58 & \cellcolor[HTML]{D5E2F3}0.65 & \cellcolor[HTML]{D1E1F4}0.67 & \cellcolor[HTML]{CCDEF5}0.70\\
E8 & Heavy noise on both images ($\sigma=0.35$) & \cellcolor[HTML]{86B6F0}1.00 & \cellcolor[HTML]{86B6F0}1.00 & \cellcolor[HTML]{F0EFEC}0.50 & \cellcolor[HTML]{DDE6F1}0.60 & \cellcolor[HTML]{86B6F0}1.00 & \cellcolor[HTML]{86B6F0}1.00 & \cellcolor[HTML]{A4C8F3}0.88 & \cellcolor[HTML]{DAE5F1}0.62 & \cellcolor[HTML]{86B6F0}1.00 & \cellcolor[HTML]{87B7F0}0.99 & \cellcolor[HTML]{8AB8F0}0.98\\
\addlinespace[2pt]
\multicolumn{13}{@{}l}{\textit{Scene}}\\
E9 & Target displaced by 6\,cm & \cellcolor[HTML]{DEE7F0}0.60 & \cellcolor[HTML]{B9D4F6}0.79 & \cellcolor[HTML]{F0EFEC}0.50 & \cellcolor[HTML]{DFE7F0}0.59 & \cellcolor[HTML]{D4E2F3}0.65 & \cellcolor[HTML]{BDD6F6}0.77 & \cellcolor[HTML]{C7DCF6}0.73 & \cellcolor[HTML]{D9E4F2}0.63 & \cellcolor[HTML]{BCD6F6}0.78 & \cellcolor[HTML]{BBD5F6}0.78 & \cellcolor[HTML]{B6D2F5}0.80\\
E10 & Another object moved next to the target & \cellcolor[HTML]{D5E2F3}0.65 & \cellcolor[HTML]{C5DBF6}0.74 & \cellcolor[HTML]{F0EFEC}0.50 & \cellcolor[HTML]{EEEEEC}0.51 & \cellcolor[HTML]{CADDF5}0.71 & \cellcolor[HTML]{C3DAF7}0.75 & \cellcolor[HTML]{D8E4F2}0.63 & \cellcolor[HTML]{E3E9EF}0.57 & \cellcolor[HTML]{C2DAF7}0.75 & \cellcolor[HTML]{CDDFF5}0.70 & \cellcolor[HTML]{D1E0F4}0.68\\
E11 & Target removed & \cellcolor[HTML]{C8DCF6}0.72 & \cellcolor[HTML]{C7DCF6}0.73 & \cellcolor[HTML]{F0EFEC}0.50 & \cellcolor[HTML]{E6EAEF}0.56 & \cellcolor[HTML]{BAD4F6}0.79 & \cellcolor[HTML]{CEDFF4}0.69 & \cellcolor[HTML]{B4D1F5}0.81 & \cellcolor[HTML]{CEDFF4}0.69 & \cellcolor[HTML]{A3C7F3}0.88 & \cellcolor[HTML]{A0C5F3}0.89 & \cellcolor[HTML]{9EC4F3}0.90\\
\addlinespace[2pt]
\multicolumn{13}{@{}l}{\textit{Robot}}\\
E12 & Arm pushed (joints offset by 0.1\,rad) & \cellcolor[HTML]{BDD7F6}0.77 & \cellcolor[HTML]{B2D0F5}0.82 & \cellcolor[HTML]{F0EFEC}0.50 & \cellcolor[HTML]{B1CFF5}0.82 & \cellcolor[HTML]{BDD6F6}0.78 & \cellcolor[HTML]{B3D1F5}0.81 & \cellcolor[HTML]{B2D0F5}0.82 & \cellcolor[HTML]{D1E0F4}0.67 & \cellcolor[HTML]{A4C8F3}0.88 & \cellcolor[HTML]{A5C8F4}0.87 & \cellcolor[HTML]{AACBF4}0.85\\
\midrule
& Mean over perturbations & & & & & & & & & \cellcolor[HTML]{B7D3F6}0.80 & \cellcolor[HTML]{B9D4F6}0.79 & \cellcolor[HTML]{BFD8F7}0.76\\
\bottomrule
\end{tabularx}

\end{table}

Tables~\ref{tab:context-start} and~\ref{tab:context-execution} (rows S1--S17 and E1--E12) report within-task AUROC for distinguishing perturbed from clean observations, initially and during execution, for the context heads, \KVctx{}, \KVdec{} and VFD.
We expect raw heads to detect unfamiliarity in their own input, fused heads to also detect unfamiliar combinations, and the whole-policy scores to respond across modalities.

Scoring only initial observations (Table~\ref{tab:context-start}), raw heads remain near chance when their input is unchanged by construction.
Unfamiliar instructions (S1--S5), workspace camera shifts (S8), wrist blackouts (S11) and reported position offsets (S16) selectively raise the corresponding raw head.
Lighting, noise, texture and scene changes (S6, S7, S10 and S12--S14) raise both camera heads, more strongly at the wrist, while joint offsets (S15) also raise the state head.
During execution, a change in one input modality can alter the robot's behaviour and lead to unfamiliar observations in other modalities.
For example, switching to another task's instruction (E2) also raises the scores of the camera and state heads.
Fused heads also respond to changes in other inputs: the fused instruction head responds to most visual and scene changes (S6--S14), and fused camera heads to strong instruction changes such as translation (S4 and S5).
The instruction of another task whose objects are absent from the scene (S17) leaves all raw heads at chance, because instruction and scene are each familiar, but raises the fused instruction and state heads, \KVctx{}, \KVdec{} and VFD.
During execution, the switch to another task's instruction (E2), which for eight of the ten tasks names objects absent from the scene, is likewise detected by \KVctx{}, \KVdec{} and VFD, while the raw instruction head stays at chance.

Both whole-policy kernel heads respond to most visual and unfamiliar instruction changes (S1--S14).
However, \KVctx{} misses reported position offsets (S16, AUROC 0.50), which the fused state head (0.69), \KVdec{} and VFD (both 0.62) partly detect.
Across the displayed perturbations of starting observations, mean AUROC is 0.86 for \KVdec{}, 0.81 for \KVctx{} and 0.80 for VFD.
On shared initial states, the paired AUROC difference between \KVdec{} and VFD has an unadjusted 95\% interval entirely above zero for 12 of 17 perturbations and entirely below zero for none.
For \KVctx{}, the corresponding interval lies entirely below zero only for the two changes to robot state (S15 and S16, Appendix Table~\ref{tab:context-paired}).
During execution, mean AUROCs are 0.79 for \KVdec{}, 0.80 for \KVctx{} and 0.76 for VFD, while none of the unadjusted intervals for either kernel head's difference from VFD lies entirely below zero.
Across perturbations of starting observations, detection strength tracks success loss, but not fully (Figure~\ref{fig:context-success}a, Spearman correlations 0.75--0.79 for \KVdec{}, \KVctx{} and VFD).
Mild image noise (S7) for example raises wrist-head scores without an observed loss of success.

\subsection{Referral under controlled within-task variation}
\label{sec:results-perturbed}

\begin{table}[t]
\centering\small
\caption{Within-task referral under perturbations to a supervisor trained with them, based on uncertainty scores of clean and perturbed starting observations (mean $\pm$ SD over three seeds). Chance level is 0 for nAUC and 0.5 for AUROC. The cross-replicate reference ranks the conditions by the supervisor's gain observed in one of the three executions of each condition and is evaluated on the other two.}
\label{tab:context-referral}
\begin{tabular}{@{}lcccc@{}}
\toprule
Score & Within-task nAUC & Global nAUC & Automated & Failure AUROC\\
\midrule
\KVctx{} & $0.40\pm0.05$ & $0.35\pm0.02$ & $0.65\pm0.22$ & $0.76\pm0.03$\\
\KVdec{} & $0.46\pm0.03$ & $0.41\pm0.02$ & $0.76\pm0.13$ & $0.79\pm0.01$\\
VFD (two policies) & $0.40\pm0.03$ & $0.39\pm0.03$ & $0.73\pm0.14$ & $0.75\pm0.00$\\
\addlinespace
Cross-replicate reference & $0.21\pm0.03$ & $0.21\pm0.03$ & $0.53\pm0.25$ & $0.62\pm0.01$\\
\bottomrule
\end{tabular}

\end{table}

\begin{figure}[t]
\centering
\includegraphics[width=\linewidth]{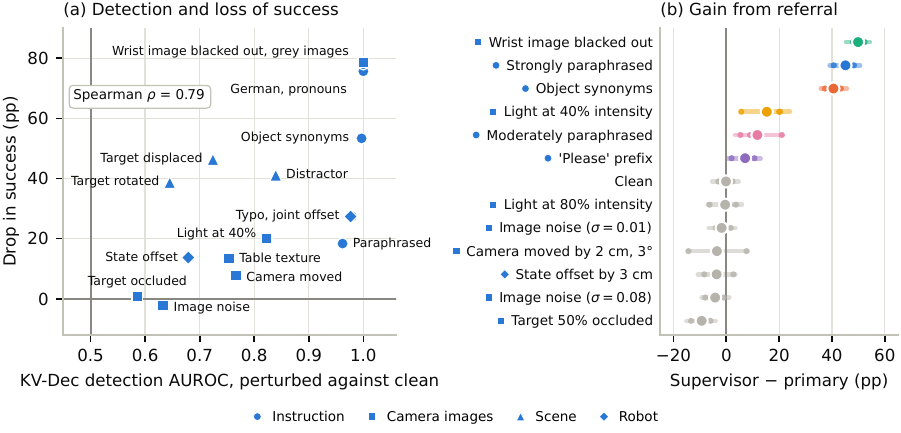}
\caption{Task success under perturbations. (a) Detection of perturbed starting observations against the loss of success they cause. Each point is one level of a perturbation family (Appendix Table~\ref{tab:perturbations}), placed by \KVdec{} AUROC of perturbed against clean starting observations and by success decline relative to clean executions from the same initial states. (b) Success of the supervisor, which is trained on these perturbations, minus that of the policy whose uncertainty is estimated, for each perturbation of the referral experiment (Table~\ref{tab:context-referral}), averaged over tasks and their ten initial states, with seed values (small dots), their mean (large dot) and the 95\% hierarchical bootstrap interval (bar). Rows are sorted by the mean, and coloured rows have an interval above zero. Symbols give the perturbed input in both panels.}
\label{fig:context-success}
\end{figure}

Standard LIBERO conditions vary little within tasks and much variation in outcomes is due to execution noise (Section~\ref{sec:results-benchmark}).
We therefore introduce input perturbations to test whether the uncertainty scores identify conditions that benefit from referral within the same task.
On ten initial states per task, we evaluate primary and supervisor on clean conditions and 12 perturbations from nine label-preserving families used in supervisor training (Appendix Table~\ref{tab:perturbations}).
The supervisor gains most under wrist blackout (50\% success versus 0\% for the primary), strong paraphrases (58\% versus 13\%) and object synonyms (68\% versus 27\%), while many perturbations are harmless (e.g., image noise, light at 80\% intensity) or hurt both policies similarly (e.g., the moved camera) (Figure~\ref{fig:context-success}b).
For target occlusion, the supervisor even loses 9 percentage points.
Across perturbed conditions, success averages 71\% for the supervisor and 58\% for the primary; on clean conditions, both reach 79\%.

Uncertainty-based referral is more effective here than on standard LIBERO conditions (Table~\ref{tab:context-referral}).
\KVdec{} reaches a within-task nAUC of $0.46\pm0.03$, compared with $0.40\pm0.05$ for \KVctx{} and $0.40\pm0.03$ for VFD.
Paired comparisons show that \KVdec{} exceeds VFD by $0.06$ $[0.03,0.10]$ in nAUC and $0.03$ $[0.02,0.05]$ in failure AUROC, whereas \KVctx{} shows no significant difference in nAUC ($0.00$ $[-0.06,0.06]$, Appendix Table~\ref{tab:context-paired}).
All three exceed the cross-replicate reference (0.21), which ranks conditions by the supervisor's gain observed in a single execution.
The kernel heads thus support within-task referral when observable perturbations expose weaknesses that the supervisor has learned to handle.

\section{Discussion}
\label{sec:discussion}

\subsection{Calibration and referral (Q1)}

Q1 asked whether a single-policy kernel head matches ensemble scores in the LIBERO-10 evaluations, calibration to task success and referral to a stronger supervisor.
The heads perform competitively in these evaluations (Sections~\ref{sec:results-calibration} and~\ref{sec:results-referral}), with advantages in particular metrics.
Both correlate with task success more linearly than VFD and GU (Table~\ref{tab:calibration} and Appendix Table~\ref{tab:paired}).
Their variance is bounded and drops to its floor for support observations by construction, so tasks with and without demonstrations form two well-separated groups.
In referral, both heads also reach a higher automated fraction than VFD and GU, supported by the intervals for \KVdec{} and, against GU, for \KVctx{} (Table~\ref{tab:referral} and Appendix Table~\ref{tab:paired}).

\subsection{Demonstration acquisition (Q2)}

Q2 asked whether the active learning protocol SAVE guided by the kernel heads acquires demonstrations as effectively as SAVE with ensemble scores.

We observe similar acquisition performance, and in our experiments every query guided by either kernel head returns a new demonstration (Figure~\ref{fig:acquisition} and Table~\ref{tab:acquisition}).
Conditioning on acquired observations assigns them low variance after refitting, encouraging subsequent queries elsewhere.
Calibration and referral only test how a score ranks fixed conditions, whereas SAVE also weights tasks by the size of their scores and refits the head after every round.
A good acquisition score must therefore also drop for what it has acquired, which exact conditioning guarantees.

\subsection{Computational cost (Q3)}

Q3 asked how much computation the kernel heads save.
The heads are much faster than the ensemble scores both to set up (in our experiments about 4 to 44 times per acquisition round) and to use (about 4 to 50 times against VFD when scoring the evaluation conditions, Section~\ref{sec:results-acquisition}).
Setting up a kernel head needs no training.
It takes one forward pass of the frozen policy over the support and one factorization of the kernel matrix, whose cost grows with the number of support inputs but not with the size of the policy.
Ensemble scores instead train at least one further policy during pre-training and fine-tuning, so their overhead grows with the policy's training cost.
Online, \KVctx{} adds only a kernel evaluation to embeddings that the policy computes anyway to act, whereas ensemble scores run every member.
\KVdec{} runs part of the action expert again, putting it in between \KVctx{} and the ensemble scores.
SmolVLA is a small VLA.
For larger policies, training a second model costs far more, so the absolute savings grow with the size of the policy.

Scoring during execution requires a support that also contains later frames of the demonstrations. With every tenth frame, in our experiments the exact factorization grows to about 10,000 rows, which is still affordable on a GPU.
For much larger supports, sparse variational GPs with inducing points~\citep{titsias2009variational,hensman2013svgp} reduce the cost from cubic to linear in the support size.
Pivoted Cholesky~\citep{harbrecht2012pivoted,burt2020convergence} bounds sparse variance inflation by a set tolerance at every support input, so acquired demonstrations still become certain.

\subsection{What the established evaluations reward (Q4)}

Q4 asked which properties of an uncertainty score the established evaluations reward on LIBERO.
They mainly reward recognizing which tasks the demonstrations cover (Section~\ref{sec:results-benchmark}).
Ensemble disagreement captures this coverage indirectly, because two policies trained on the same data agree where data are dense.
A kernel head measures it directly in the policy's own embedding space, as a label-free count of nearby demonstrations.
Both therefore rank tasks alike, which explains the parity found for Q1.

High task-level calibration is necessary for prioritizing tasks but is a weak test.
In the evaluation protocol of \citet{romer2026uqvla} that we reproduce, it mainly reflects how many demonstrations each task has, so a plain demonstration count reaches it as well (Table~\ref{tab:calibration}).
Recognizing this coverage is still useful, because demonstration counts are unavailable for new tasks and scenes in deployment.
Ranking situations within a task is a separate question, and on LIBERO's standard initial conditions it was largely inconclusive.
LIBERO task conditions differ little and reproducible within-task differences in outcome are small compared to outcome variations during the policy's execution (Figure~\ref{fig:benchmark}a).

We suggest that evaluations of uncertainty for adaptive robot learning report a coverage reference such as the demonstration count, separate global from within-task metrics, quantify execution noise with repeated executions, and include within-task variation that is known to matter, as in LIBERO-PRO, LIBERO-Plus~\citep{zhou2025liberopro,fei2025liberoplus} and our perturbed conditions.

\subsection{Attributing unfamiliarity to input modalities (Q5)}

We find that raw context heads detect unfamiliar perturbations of the inputs they read (Table~\ref{tab:context-start} and Table~\ref{tab:context-execution}), while their insensitivity to changes confined to other inputs follows by construction.
Several heads can share the same frozen policy, providing separate familiarity estimates that could guide targeted assistance, such as checking the wrist camera or rephrasing an instruction.
The heads on fused embeddings detect unfamiliar combinations of individually familiar inputs. For example, an instruction naming objects absent from the scene leaves all raw heads at chance but raises the fused instruction and state heads (S17 in Table~\ref{tab:context-start}).
During execution, changes in one modality can also alter the robot's behaviour and produce subsequent unfamiliar observations in others, as shown by the switch to another known but unrelated instruction in E2 (Table~\ref{tab:context-execution}).
Separate context heads may also reveal confined changes that the whole-policy scores miss.
\KVctx{}, \KVdec{} and VFD do not or only partly detect the reported position offset, whereas the raw state head responds perfectly (S16).

\subsection{When unfamiliarity supports within-task referral (Q5)}

We find that within-task referral is possible when meaningful variation in the task conditions exists. \KVdec{} improves on VFD, whereas \KVctx{} performs comparably (Table~\ref{tab:context-referral} and Appendix Table~\ref{tab:context-paired}).
Referral helps where the supervisor is more likely to succeed.
Here, the largest gains occur under wrist blackout, strong paraphrases and object synonyms, which the supervisor was successfully trained to handle (Figure~\ref{fig:context-success}b).
If both policies fail or the primary already copes, referral offers no expected benefit.
Detecting unfamiliarity alone therefore does not ensure useful referral.

Execution variability makes gains observed in one execution noisy predictors of later gains.
The input-based rankings of the whole-policy scores support referral considerably more effectively than a single-execution estimate, which ranks conditions by the supervisor's gain in one execution and evaluates on others (Table~\ref{tab:context-referral}).

\subsection{Further findings}

\paragraph{Unfamiliarity is not harm.}
Our kernel heads estimate epistemic uncertainty under a GP model on the policy's embeddings.
Equation~\eqref{eq:kv} uses no outputs and shrinks only when data are added.
They therefore measure how well the support covers an input in the policy's embedding space, not whether the policy will fail on it.
If the embeddings keep unfamiliar inputs apart from familiar ones, this coverage reflects how unfamiliar the input is.
In general, detection strength correlates quite strongly with the loss of success (Figure~\ref{fig:context-success}a).
However, some perturbations that \KVdec{} detects almost perfectly cost anywhere from less than 20 points of success to complete failure, while scene changes that it detects only moderately cost about 40 points (Figure~\ref{fig:context-success}a).

\paragraph{Single-policy entropy can correlate with coverage.}
ACE, the action-chunk entropy of a single policy, cannot capture epistemic uncertainty by design, yet for the policies of Table~\ref{tab:calibration} it correlates with training-data coverage across tasks.
On tasks without demonstrations, the policy has to extrapolate, and if its aleatoric estimate of action spread is on average wider there, ACE follows coverage without representing epistemic uncertainty.
Indeed, before acquisition, the policy's sampled actions spread more widely on the seven tasks without demonstrations, so ACE ranks tasks by their demonstration count (Spearman $0.64$).
After ten rounds, when every task has some demonstrations, this correlation already drops to $0.21$, while VFD, for example, still reaches $0.82$.
Across all nine scores, the correlation of a score with the demonstration count closely matches its task-level calibration (Pearson above $0.99$ before and $0.94$ after acquisition), which suggests that the moderate calibration of ACE in Table~\ref{tab:calibration} mainly reflects its correlation with coverage.
As discussed in Section~\ref{sec:related}, such agreement is indirect.

\subsection{Limitations}

We study one policy architecture, SmolVLA, on one simulated benchmark with ten target tasks and three training seeds.
Generalization to other architectures, actual robots, and real-world conditions remains untested. The main bootstrap intervals characterize uncertainty for the evaluated tasks rather than variation across unseen tasks.
Three seeds estimate training variability only coarsely, and three executions per condition leave success rates and supervisor gains noisy.
Inconclusive comparisons therefore do not establish equivalence.

The kernel variance measures demonstration coverage under a chosen representation and GP prior.
Its interpretation as the policy's epistemic uncertainty depends on whether embedding distances preserve distinctions relevant to action selection.
We do not enforce this property during training.
Feature collapse, token averaging, or dimensionality reduction can hide relevant differences.
Low variance at a support observation does not establish that the policy has learned to act successfully there, while unfamiliar inputs can be harmless.

Calibration and referral in the matched study (based on \citet{romer2026uqvla}) use only two checkpoints of the random-acquisition arm, due to computational limitations.
Their task-level calibration measures correlation with success rather than calibrated failure probabilities.
Our SAVE/VFD reproduction also remains below the final success reported by \citet{romer2026uqvla}. Although this may plausibly be explained by variability across training runs, this difference limits the acquisition comparison.
Furthermore, acquisition selects from a fixed pool of recorded demonstrations, and the inherited eligibility rule permits repeated requests.
Since the kernel heads acquire more unique demonstrations, the results do not isolate better selection among new demonstrations from avoidance of repeated queries.

The perturbation experiments use separate policies trained on all ten target tasks and support observations sampled throughout their demonstrations.
They therefore extend the evaluation under a different training and support regime.
The perturbations are deliberately constructed, and the observed relationship between detection and loss of success is specific to their selection.
The referral supervisor is trained on the same label-preserving perturbation families used in evaluation.
Useful referral under unfamiliar failure mechanisms remains untested.

Referral evaluates delegation before execution to an automated supervisor proxy starting independently from the same initial state.
It does not test human assistance, handover during execution, recovery, or intervention costs.
The execution experiments establish sensitivity to scheduled perturbations using two initial states per task and a fixed post-perturbation window.
They do not establish warning before failure, suitable episode-level false-alarm rates, or improved outcomes from responding to an alarm.

Context-specific heads identify unfamiliarity in individual representations, but do not establish which input caused a failure or which corrective intervention would help.
Raw-head selectivity partly follows by construction, while fused responses depend on cross-modal mixing and subsequent behaviour.
Targeted assistance based on these signals remains a proposed application.

Finally, head configurations were chosen through limited exploratory studies rather than a systematic sensitivity analysis.
Comparisons with equally inexpensive distance or density scores would help establish the contribution of exact GP conditioning.
The reported computational savings also apply to the tested policies, hardware, and support sizes. Exact conditioning requires cubic factorization time and quadratic storage, and policy updates require extracting support embeddings again.
Supports beyond approximately 10,000 observations remain unevaluated.
Sparse approximations could extend this range, but their effects on ranking quality, acquisition behaviour, and computational savings remain untested.

\section{Conclusion}
\label{sec:conclusion}

We introduced post-hoc GP kernel-variance heads that estimate epistemic uncertainty from the embeddings of a single VLA policy.
Fitting requires no regression targets or gradient updates and leaves the policy unchanged.
On SmolVLA and LIBERO-10, the heads performed competitively with two-policy ensemble scores in task-level calibration, demonstration acquisition, and supervisor referral, while substantially reducing computational overhead.
For SmolVLA, \KVctx{} reduced uncertainty-scoring time by up to approximately a factor of 50 and setup time per acquisition round by up to approximately a factor of 44 compared with ensemble scores.

Our analysis also clarifies what these evaluations reward.
In the evaluation protocol of \citet{romer2026uqvla}, task-level calibration and global referral primarily reflect differences in demonstration coverage between tasks.
They provide weaker evidence about ranking situations within a task, where execution variability obscures reproducible differences in outcomes.
Evaluating uncertainty for adaptive robot learning therefore benefits from coverage references, repeated executions, and separate reporting of global and within-task performance.

Controlled perturbations reveal further capabilities of the kernel-variance heads.
Context-specific heads identify unfamiliar input modalities, while fused heads can detect unfamiliar combinations of individually familiar inputs.
Under these perturbations, our \KVdec{} improves within-task referral over VFD to a supervisor trained to handle them.
However, unfamiliarity supports useful referral only where the supervisor has a performance advantage. Detecting unfamiliarity and determining the benefit of assistance remain distinct questions.

Future work should test these findings across other VLA architectures and physical robots, establish the contribution of GP conditioning relative to simpler distance and density scores, and evaluate whether context-specific unfamiliarity can guide effective assistance during execution. Such evaluations should assess intervention timing, task success, and assistance costs alongside detection quality.

\bibliographystyle{unsrtnat}
\bibliography{references}
\clearpage
\appendix
\section{Head configurations and implementation}
\label{app:implementation}
Table~\ref{tab:heads} summarizes the heads of Section~\ref{sec:kv}.
We selected the head configurations informally from limited exploratory experiments, retaining variants that performed well rather than following a systematic search.
Features are extracted with the policy in evaluation mode, one observation per call, so that the fixed initial noise of \KVdec{} does not depend on batch position.
\KVctx{} and the context-specific heads remove duplicate support inputs before the fit, whereas \KVdec{} keeps them.

\begin{table}[ht]
\centering\small
\caption{Kernel-variance heads in the main experiments. All heads use exact GP conditioning on their support observations and need no regression targets.}
\label{tab:heads}
\begin{tabularx}{\linewidth}{@{}l>{\raggedright\arraybackslash}X>{\raggedright\arraybackslash}X@{}}
\toprule
Head & Embedding & Geometry and kernel\\
\midrule
\KVctx{} & Final prefix states: mean per camera and over instruction tokens, plus 8 state coordinates (2,888) & Standardize, PCA512, Ledoit--Wolf whitening, $k$-NN bandwidth, $\lambda=10^{-3}$\\
\KVdec{} & Action-expert final hidden state at flow step 9 of 10, mean over 50 action tokens (480) & Standardize, full Ledoit--Wolf whitening, median-distance bandwidth, $\lambda=10^{-2}$\\
Context heads & Raw: mean token embedding of the workspace camera, wrist camera and instruction (960 each), 8 state values. Fused: final prefix states of the same groups and of the state token (960 each) & As \KVctx{}, scores compared as percentiles among clean calibration observations\\
\bottomrule
\end{tabularx}
\end{table}

\section{Additional calibration and referral results}
\label{app:additional}

This appendix extends the task-level calibration and supervisor referral of the matched SAVE study (Sections~\ref{sec:results-calibration} and~\ref{sec:results-referral}), which score the primaries of the Random arm at rounds 0 and 10.
Table~\ref{tab:per-round} separates the two common-policy checkpoints, and Table~\ref{tab:paired} gives paired differences between the kernel heads and the ensemble scores with hierarchical bootstrap intervals.
A difference whose interval excludes zero is supported for these tasks and this evaluation population. With three seeds, variation between training runs is estimated only coarsely.

\begin{table}[ht]
\centering\scriptsize
\setlength{\tabcolsep}{3pt}
\caption{Calibration and referral at each common-policy checkpoint (mean $\pm$ SD over three seeds).}
\label{tab:per-round}
\resizebox{\linewidth}{!}{\begin{tabular}{@{}lcccccccc@{}}
\toprule
& \multicolumn{4}{c}{Round 0} & \multicolumn{4}{c}{Round 10}\\
\cmidrule(lr){2-5}\cmidrule(l){6-9}
Score & $-\rho_{\rm S}$ & $-r_{\rm P}$ & Global & Within & $-\rho_{\rm S}$ & $-r_{\rm P}$ & Global & Within\\
\midrule
VFD & $0.83\pm0.05$ & $0.75\pm0.05$ & $0.65\pm0.05$ & $0.08\pm0.05$ & $0.66\pm0.27$ & $0.54\pm0.29$ & $0.31\pm0.09$ & $0.07\pm0.05$\\
GU & $0.81\pm0.04$ & $0.89\pm0.02$ & $0.65\pm0.03$ & $0.05\pm0.03$ & $0.50\pm0.40$ & $0.47\pm0.40$ & $0.20\pm0.10$ & $0.06\pm0.03$\\
Action-L2 & $0.81\pm0.04$ & $0.66\pm0.06$ & $0.58\pm0.05$ & $0.06\pm0.06$ & $0.32\pm0.14$ & $0.34\pm0.24$ & $0.13\pm0.07$ & $0.06\pm0.04$\\
ACE & $0.72\pm0.11$ & $0.67\pm0.05$ & $0.43\pm0.05$ & $0.01\pm0.08$ & $0.15\pm0.28$ & $0.20\pm0.26$ & $0.15\pm0.09$ & $0.06\pm0.10$\\
DECU & $0.50\pm0.39$ & $0.44\pm0.11$ & $0.26\pm0.19$ & $-0.03\pm0.08$ & $0.24\pm0.42$ & $0.36\pm0.26$ & $0.14\pm0.08$ & $0.03\pm0.01$\\
VLM entropy & $-0.21\pm0.52$ & $-0.04\pm0.34$ & $-0.06\pm0.17$ & $0.03\pm0.07$ & $0.03\pm0.22$ & $-0.02\pm0.27$ & $-0.05\pm0.15$ & $0.02\pm0.03$\\
VLM perplexity & $-0.19\pm0.30$ & $-0.16\pm0.32$ & $-0.18\pm0.11$ & $-0.04\pm0.03$ & $-0.08\pm0.19$ & $0.15\pm0.23$ & $-0.08\pm0.06$ & $-0.10\pm0.01$\\
\midrule
\KVctx{} & $0.84\pm0.04$ & $0.94\pm0.02$ & $0.63\pm0.01$ & $0.02\pm0.08$ & $0.69\pm0.16$ & $0.70\pm0.19$ & $0.38\pm0.08$ & $0.05\pm0.12$\\
\KVdec{} & $0.86\pm0.04$ & $0.96\pm0.01$ & $0.63\pm0.02$ & $0.02\pm0.04$ & $0.64\pm0.35$ & $0.63\pm0.28$ & $0.39\pm0.12$ & $0.11\pm0.03$\\
\bottomrule
\end{tabular}
}
\end{table}

\begin{table}[ht]
\centering\scriptsize
\setlength{\tabcolsep}{3pt}
\caption{Paired differences averaged over rounds 0 and 10, with 95\% hierarchical bootstrap intervals (2,000 draws over conditions within tasks and over seeds, shared across scores).}
\label{tab:paired}
\resizebox{\linewidth}{!}{\begin{tabular}{@{}lccccc@{}}
\toprule
Difference & $-\rho_{\rm S}$ & $-r_{\rm P}$ & Global nAUC & Within nAUC & Automated fraction\\
\midrule
\KVctx{} $-$ VFD & $+0.020$ $[-0.021, +0.090]$ & $+0.176$ $[+0.106, +0.254]$ & $+0.022$ $[-0.019, +0.063]$ & $-0.041$ $[-0.142, +0.062]$ & $+0.044$ $[-0.014, +0.082]$\\
\KVctx{} $-$ GU & $+0.108$ $[+0.003, +0.225]$ & $+0.139$ $[+0.045, +0.244]$ & $+0.082$ $[+0.039, +0.122]$ & $-0.022$ $[-0.143, +0.092]$ & $+0.068$ $[+0.025, +0.117]$\\
\KVdec{} $-$ VFD & $+0.003$ $[-0.036, +0.059]$ & $+0.152$ $[+0.112, +0.187]$ & $+0.028$ $[-0.005, +0.059]$ & $-0.011$ $[-0.079, +0.060]$ & $+0.084$ $[+0.030, +0.111]$\\
\KVdec{} $-$ GU & $+0.091$ $[+0.030, +0.129]$ & $+0.115$ $[+0.051, +0.179]$ & $+0.087$ $[+0.048, +0.124]$ & $+0.008$ $[-0.070, +0.084]$ & $+0.108$ $[+0.068, +0.149]$\\
\bottomrule
\end{tabular}
}
\end{table}

\section{Context-head experiments}
\label{app:context}
This appendix complements the context-head experiments (Sections~\ref{sec:results-rollout} and~\ref{sec:results-perturbed}) with the perturbation library (Table~\ref{tab:perturbations}) and the paired comparisons of the paper heads with VFD (Table~\ref{tab:context-paired}).

\begin{table}[ht]
\centering\scriptsize
\caption{Perturbation library, with the experiments that use each family. Label-preserving: the demonstrated actions remain correct, so the supervisor can be trained on the perturbation. P3 and P5: two of the released paraphrases of each instruction, which range from close (P2) to distant (P5).}
\label{tab:perturbations}
\begin{tabularx}{\linewidth}{@{}>{\raggedright\arraybackslash}p{2.1cm}>{\raggedright\arraybackslash}Xcccc@{}}
\toprule
Input & Perturbation and levels & \shortstack{Starting\\observ.} & \shortstack{During\\execution} & Referral & \shortstack{Label-\\preserving}\\
\midrule
instruction & graded paraphrase: P3; P5 & $\checkmark$ & $\checkmark$ & $\checkmark$ & yes\\
instruction & object synonym: all object words replaced & $\checkmark$ &  & $\checkmark$ & yes\\
instruction & surface variation: 'please ' prefix; one typo in a non-object word & $\checkmark$ &  & $\checkmark$ & yes\\
instruction & underspecified: object nouns replaced by pronouns ('put it in the basket') & $\checkmark$ &  &  & no\\
instruction & blatant: negated ('do not ...'); German translation & $\checkmark$ & $\checkmark$ &  & no\\
vision (both cameras) & lighting change: intensity x0.8; x0.4 & $\checkmark$ & $\checkmark$ & $\checkmark$ & yes\\
vision (both cameras) & sensor degradation: Gaussian noise sigma 0.01; sigma 0.08 & $\checkmark$ &  & $\checkmark$ & yes\\
vision (workspace camera) & camera pose shift ('bumped camera'): 2 cm + 3 deg; 4 cm + 6 deg & $\checkmark$ & $\checkmark$ & $\checkmark$ & yes\\
vision (workspace camera) & target partly occluded: 50\%; 75\% & $\checkmark$ & $\checkmark$ & $\checkmark$ & yes\\
vision (scene appearance) & texture or colour change: table texture swap & $\checkmark$ &  &  & yes\\
vision (wrist camera) & wrist camera fault: blackout; frozen frame (during execution: repeat the frame at onset) & $\checkmark$ & $\checkmark$ & $\checkmark$ & yes\\
vision & blatant image corruption: Gaussian noise sigma 0.35; uniform grey image & $\checkmark$ & $\checkmark$ &  & no\\
scene & target displaced outside its training region: 6 cm; 10 cm & $\checkmark$ & $\checkmark$ &  & no\\
scene & extra distractor: next to the target ($<$ 5 cm) & $\checkmark$ & $\checkmark$ &  & no\\
scene & target rotated: 90 deg & $\checkmark$ &  &  & no\\
scene & objects vanish or scrambled: vanish target &  & $\checkmark$ &  & no\\
robot state & initial joint offset (robot not at home pose): 0.1 rad & $\checkmark$ &  &  & no\\
robot state & proprioception bias: 3 cm & $\checkmark$ &  & $\checkmark$ & yes\\
robot state & push during execution: 0.1 rad &  & $\checkmark$ &  & no\\
instruction $\times$ scene & other task's instruction: same scene: tasks 0 and 1; other scene: an instruction naming objects that are absent & $\checkmark$ & $\checkmark$ &  & no\\
\bottomrule
\end{tabularx}

\end{table}

\begin{table}[ht]
\centering\small
\caption{Paired differences between the paper heads and VFD in the context-head experiments: mean over three seeds with 95\% hierarchical bootstrap intervals for referral, and the number of perturbations (Table~\ref{tab:context-start}) or events (Table~\ref{tab:context-execution}) for which the paper head's AUROC is significantly higher, not significantly different or significantly lower than VFD's. On starting observations, the comparison uses the ten initial states per task on which VFD is computed. Intervals are unadjusted for multiple comparisons, and perturbation-wise comparisons are exploratory.}
\label{tab:context-paired}
\resizebox{\linewidth}{!}{\begin{tabular}{@{}llcccc@{}}
\toprule
Experiment & Difference & \multicolumn{4}{c}{Mean over seeds, 95\,\% hierarchical interval}\\
\cmidrule(l){3-6}
Referral (Table~\ref{tab:context-referral}) & & Within-task nAUC & Global nAUC & Automated & Failure AUROC\\
\midrule
& \KVctx{} $-$ VFD & $0.00$ {\scriptsize$[-0.06, +0.06]$} & $-0.05$ {\scriptsize$[-0.10, +0.01]$} & $-0.08$ {\scriptsize$[-0.21, +0.02]$} & $+0.01$ {\scriptsize$[-0.02, +0.04]$}\\
& \KVdec{} $-$ VFD & $+0.06$ {\scriptsize$[+0.03, +0.10]$} & $+0.02$ {\scriptsize$[-0.01, +0.06]$} & $+0.03$ {\scriptsize$[-0.02, +0.10]$} & $+0.03$ {\scriptsize$[+0.02, +0.05]$}\\
\midrule
Detection AUROC & & Higher & Not different & Lower & Median difference\\
\midrule
Starting observations (Table~\ref{tab:context-start}) & \KVctx{} $-$ VFD & 4 & 11 & 2 & $+0.01$\\
 & \KVdec{} $-$ VFD & 12 & 5 & 0 & $+0.04$\\
During execution (Table~\ref{tab:context-execution}) & \KVctx{} $-$ VFD & 4 & 8 & 0 & $+0.02$\\
 & \KVdec{} $-$ VFD & 3 & 9 & 0 & $+0.02$\\
\bottomrule
\end{tabular}
}
\end{table}

\end{document}